\documentclass{fairmeta}

\usepackage{amsmath}
\usepackage{amsfonts}
\usepackage{nicefrac}
\usepackage{wrapfig}
\usepackage{pifont}
\usepackage{xspace}
\usepackage{xurl}
\usepackage{url}
\usepackage{fancyvrb}
\usepackage[subtle, mathdisplays=tight, charwidths=tight, leading=normal]{savetrees}

\usepackage{enumitem}

\newcommand{\cmark}{\textcolor{green!60!black}{\ding{51}}}
\newcommand{\xmark}{\textcolor{red!75!black}{\ding{55}}}
\newcommand{\pmark}{\textit{p}}

\definecolor{myexampleframe}{HTML}{3B7597}
\definecolor{myreportframe}{HTML}{9A8678}

\tcbset{
  myexample/.style={
    colback=white,
    colframe=myexampleframe,
    colbacktitle=myexampleframe,
    coltitle=white,
    fonttitle=\bfseries,
    arc=4mm,
    boxrule=1pt,
    left=3mm, right=3mm, top=1.5mm, bottom=1.5mm,
    title={#1}
  },
  myreport/.style={
    colback=white,
    colframe=myreportframe,
    colbacktitle=myreportframe,
    coltitle=white,
    fonttitle=\bfseries,
    arc=4mm,
    boxrule=1pt,
    left=3mm, right=3mm, top=1.5mm, bottom=1.5mm,
    title={#1}
  }
}

\newif\ifcomments
\commentstrue
\ifcomments
    \providecommand{\ion}[1]{{\color{blue}{/* ion: #1 */}}}
\else
    \providecommand{\ion}[1]{}
\fi

\newcommand{\sys}{AstraFlow\xspace}

\title{
{\fontsize{16pt}{10pt}\selectfont
    \sys: Dataflow-Oriented Reinforcement Learning for Agentic LLMs
}
}

\author{
Haizhong Zheng$^{1}$,
Yizhuo Di$^{1}$,
Jiahui Wang$^{1}$,
Shuowei Jin$^{2}$,
Xueshen Liu$^{2}$,
Yongji Wu$^{3}$, 
\\
Z. Morley Mao$^{2}$,
Ion Stoica$^{3}$,
Jiawei Zhao$^{4}$,
Beidi Chen$^{1}$ \\
$^1$Carnegie Mellon University \quad
$^2$University of Michigan \quad
$^3$UC Berkeley \quad
$^4$Meta \\
}

\metadata[Github]{\url{https://github.com/Infini-AI-Lab/astraflow}}
\metadata[Website]{\url{https://infini-ai-lab.github.io/astraflow}}

\abstract{
Reinforcement learning~(RL) is increasingly used to improve the reasoning, coding, and tool-use capabilities of large language models, but agentic RL remains prohibitively expensive.
Scaling RL to agentic LLMs requires supporting complex workloads, including multi-policy collaborative training, while efficiently using elastic, heterogeneous, and cross-region compute resources.
Existing LLM RL systems support some of these capabilities, but each new extension often requires dedicated system engineering. This burden arises from trainer-centered control architectures and the lack of principled abstractions for RL system components. To address these limitations, we propose \sys, a dataflow-oriented RL system that replaces conventional trainer-centered control with principled component abstractions. In \sys, rollout services, dataflow management, and training are decoupled into autonomous components, enabling the system to natively support complex multi-policy agentic RL workloads and efficiently exploit diverse compute resources.
We evaluate \sys across math, code, search, and AgentBench workloads, showing that the same system supports multi-policy training, elastic scaling, heterogeneous cross-region execution, and composable data algorithms without system-level code changes. In multi-policy collaborative training, \sys achieves comparable or better accuracy than existing RL systems while speeding up training time by 2.7$\times$.
}

\begin{document}

\maketitle

\begin{figure}[h]
\centering
\includegraphics[width=\linewidth]{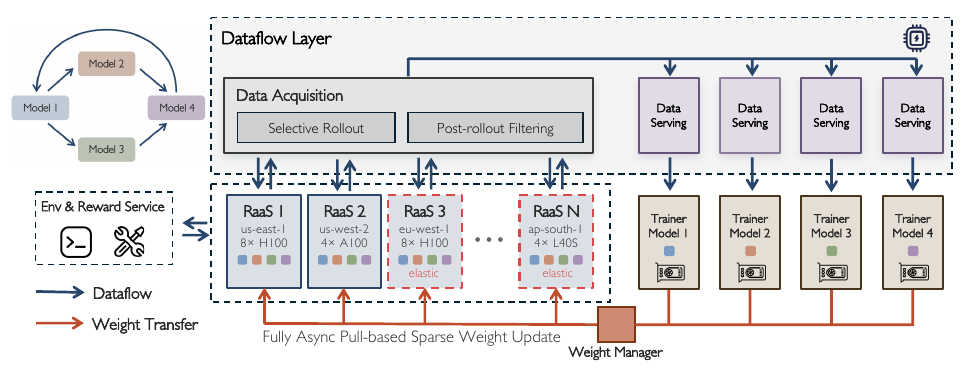}
\caption{Overview of the \sys architecture. A dataflow-oriented RL framework natively supports multi-policy collaborative training, elastic rollout, heterogeneous and cross-region rollout, and substitutable Rollout-as-a-Service~(RaaS) and Trainer.}
\label{fig:astraflow-overview}
\end{figure}

\section{Introduction}
\label{sec:introduction}

% P1: Research problem and why it matters
Large language models~(LLMs) are rapidly moving beyond standalone use into complex agentic systems, including coding agents~\citep{wei2025swerl, jimenez2023swe}, search agents~\citep{zheng2025deepresearcher, gao2025beyond}, and multi-agent workflows~\citep{cemri2025multi, jin2025search}.
In these settings, reinforcement learning~(RL) has emerged as a key technique for improving reasoning and tool-using capabilities~\citep{ouyang2022training, deepseekai2025deepseekr1incentivizingreasoningcapability, yu2025dapo, team2025kimi, shao2024deepseekmath}.
Yet scaling RL for agentic systems remains challenging. It must accommodate the complexity of agentic workloads, including dynamic execution and multi-policy coordination, as well as the diversity of underlying compute environments, such as elastic and heterogeneous computing.
This underscores the need for a general RL infrastructure that unifies agent execution, training, and resource management under a flexible and scalable system design.

Despite this need, existing systems~\citep{sheng2024hybridflow, wu2025rlboost, cao2025skyrl, hilton2025art} remain constrained by rigid designs that limit their scalability and extensibility.
\textbf{First}, existing LLM RL systems~\citep{sheng2024hybridflow} are primarily designed for single-policy training.
Their trainer-centered control logic coordinates rollout scheduling, data movement, policy optimization, and weight synchronization, making them rigid for multi-policy agentic RL, where collaborative training requires coordinating multiple policies and their interactions.
\textbf{Second}, SOTA multi-agent systems~\citep{han2024llm, wu2024autogen} primarily focus on serving.
They are designed to execute complex agent workflows efficiently, but lack the training-time coordination needed to collaboratively optimize multiple policies.
% Second, although multi-agent serving systems~\cite{han2024llm, wu2024autogen} can execute complex agent workflows, they are built for deployment rather than RL training and do not provide the training-time coordination needed to collaboratively optimize multiple policies.
% Finally, recent solutions to support capacity, like multi-policy training and utilizing diverse computation resources, are all ad-hoc patches on the existing system.
\textbf{Finally}, recent systems can be engineered to add individual capabilities such as multi-policy training~\cite{zhao2025stronger, feng2026dr}, elastic rollout~\cite{wu2025rlboost}, or heterogeneous rollout~\cite{yan2025areal, he2025hetrl}.
Although these systems can support such capabilities, they do so through ad-hoc patches that require feature-specific system engineering on the existing design.
% However, without principled abstraction boundaries among rollout execution, dataflow management, training, and weight transfer, composing these capabilities or integrating new engines and data algorithms still requires feature-specific system engineering.
Table~\ref{tab:system_comparison} summarizes how existing systems support each capability but generally lack the abstractions needed to support and compose them natively.

\begin{table}[h]
    \vspace{0.2cm}
    \centering
    \caption{System-level comparison of LLM RL frameworks. \cmark{} denotes full support, \xmark{} no support, and \pmark{} partial support.}
    \label{tab:system_comparison}
    \resizebox{\textwidth}{!}{%
    \begin{tabular}{lccccccc}
    \toprule
    \textbf{Property} & \textbf{\sys} & \textbf{AReaL} & \textbf{SLIME} & \textbf{verl}\textsuperscript{\dag} & \textbf{RLBoost} & \textbf{Dr.MAS~(verl)} & \textbf{prime-rl} \\
    \midrule
    Multi-policy collaborative training             & \cmark & \xmark & \xmark & \xmark & \xmark & \cmark & \xmark \\
    Substitutable trainer and rollout service     & \cmark & \xmark & \xmark & \xmark & \xmark & \xmark & \pmark \\
    Modular data algorithm interfaces            & \cmark & \xmark & \xmark & \xmark & \xmark & \xmark & \xmark \\
    Fully asynchronous training                         & \cmark & \cmark & \cmark & \cmark & \xmark & \xmark & \cmark \\
    Disaggregated rollout-training architecture         & \cmark & \cmark & \cmark & \cmark & \cmark & \xmark & \cmark \\
    Runtime elastic rollout scaling                     & \cmark & \xmark & \xmark & \xmark & \cmark & \xmark & \pmark \\
    Cross-region / heterogeneous rollout                & \cmark & \xmark & \xmark & \xmark & \xmark & \xmark & \pmark \\
    \bottomrule
    \end{tabular}%
    }
    \begin{minipage}{\textwidth}
    \footnotesize
    \textsuperscript{\dag}The original verl paper does not present fully asynchronous or disaggregated architecture; these entries reflect support later added in the open-source verl repository.
    \end{minipage}
    \vspace{0.2cm}
\end{table}

% P3: What an ideal solution looks like
Building an ideal argentic RL system for LLMs requires rethinking several assumptions built into conventional LLM RL systems.
First, such a system should treat multiple trainable policies and trainers as first-class components, rather than assuming a single-model, single-trainer control workflow.
Second, it should also move beyond the assumption of a fixed compute environment, allowing workloads to seamlessly run across heterogeneous, cross-region, or elastic resources through clean execution interfaces.
Third, it should move beyond tightly coupled implementations by exposing simple interfaces between rollout engines, trainers, and data algorithms, so that each component can be replaced or extended independently.

Our key insight is that the limitation of existing RL systems comes from a single trainer-centered control loop and the lack of principled abstractions among RL components.
To address this, we propose \textbf{\emph{\sys}}, a dataflow-oriented RL training system for agentic LLMs. As shown in Fig.~\ref{fig:astraflow-overview}, \sys consists of three components: a dataflow layer, Rollout-as-a-Service~(RaaS), and trainers.
\textbf{1) Dataflow layer.}
The dataflow layer coordinates rollout, training, and data-processing components through shared data, rather than centralized trainer control~(Section~\ref{sec:alt-dataflow}). This enables autonomous rollout services and trainers to compose naturally, supports multi-policy collaborative training, and expresses policies such as curriculum scheduling, replay, data mixing, filtering, sampling, and staleness correction as dataflow policies.
\textbf{2) Rollout-as-a-Service.}
RaaS decouples trajectory generation from policy optimization through rollout interfaces~(Section~\ref{sec:alt-raas}). This allows users to plug in optimized agent inference engines or specialized rollout backends without modifying trainers or system orchestration. It also enables rollout components to scale independently across heterogeneous, cross-region, and elastic compute resources.
\textbf{3) Trainers.}
Trainers consume data from the dataflow layer and publish updated weights back to the system~(Section~\ref{sec:alt-trainer-weight}). Since they no longer directly control rollout scheduling, data movement, or rollout-runtime details, trainers become independently replaceable. This makes it easy to integrate fault-tolerant trainers, specialized optimizers, or multiple trainers for multi-policy learning without changing the rest of the system.

In the evaluation part, we demonstrate the flexibility of \sys from three perspectives: multi-policy collaborative RL, system flexibility, and data algorithm flexibility.
For multi-policy collaborative RL, we evaluate \sys on three multi-policy workflows, achieving comparable or better accuracy than the existing multi-agent RL system while delivering an up to 2.7$\times$ speedup in training.
Also, to the best of our knowledge, even without any multi-agent-specific system modifications, \sys is the first fully asynchronous multi-policy collaborative RL framework.
For system flexibility, we first show that, without requiring any code changes, rollout auto-scaling can be achieved with an agentic maintainer.
Then we show that \sys natively supports heterogeneous and cross-region training without feature-specific engineering.
For data flexibility, we demonstrate the flexibility of the dataflow-layer abstraction by integrating and composing data algorithms, including dynamic sampling~\citep{yu2025dapo}, GRESO~\citep{zheng2025act}, and buffer replay.
Together, we demonstrate that, thanks to the dataflow-oriented RL design, \sys natively supports multi-policy collaborative training, diverse compute environments, and composable data algorithms without feature-specific system code.

% \sys implements GRESO~\citep{zheng2025act}, a selective prompting algorithm, buffer replay, and dynamic sampling~\citep{yu2025dapo} as modular data policies in the dataflow layer.

\section{Related Work}
\label{sec:background}

\subsection{RL for Agentic LLMs}
\label{sec:rw-agentic-rl}

RL has become a central post-training technique for improving LLM reasoning, code generation, and tool-use capabilities~\citep{ouyang2022training, shao2024deepseekmath, deepseekai2025deepseekr1incentivizingreasoningcapability, yu2025dapo, team2025kimi}.
Many efforts improve the performance, stability, and efficiency of RL itself, including better policy-optimization objectives~\cite{schulman2017proximal, yue2025vapo, zheng2025group}, reward design~\citep{wang2026rlanything}, off-policy or asynchronous training~\cite{zheng2025prosperity, noukhovitch2024asynchronous}, and data-centric algorithms~\cite{sunimproving, zheng2025act, xia2024less, xu2025not} that decide which prompts for sampling, which trajectories to keep, and which batches to train on.
At the same time, RL workloads are expanding from single-turn reasoning to more complex agentic settings~\citep{cao2025skyrl, wang2026marti, zhang2025agentrlscalingagenticreinforcement} such as software-engineering agents~\citep{wei2025swerl, jimenez2023swe}, search tasks~\citep{zheng2025deepresearcher}, and os environment interaction workflows~\citep{lai2025computerrl}.
These workloads introduce heterogeneous rollouts with variable lengths, tool feedback, intermediate artifacts, and data-policy interventions throughout training.
Recent multi-agent RL workloads~\citep{zhao2025stronger, feng2026dr} require collaboration among multiple trainable policies, further complicating training orchestration.
Together, these trends make LLM RL workloads more complex and expensive, creating a need for better system support to run them efficiently on hardware resources.

\subsection{LLM RL Training Frameworks}
\label{sec:bg-rl-systems}

A typical RL training pipeline for LLMs has two major stages: \emph{1) rollout}, where inference engines generate trajectories and rewards from the current policy, and \emph{2) training}, where a trainer consumes rollout data and updates the policy.
Existing LLM RL systems~\citep{primerl, sheng2024hybridflow, hilton2025art, wu2025rlboost, cao2025skyrl, hilton2025art, shen2024nemo, hu2024openrlhf, mei2024realhf, zhong2025optimizing, han2025asyncflow, he2025history, zhong2025streamrl} mainly organize these two stages in two ways.
\textbf{Colocated synchronous systems} such as verl~\citep{sheng2024hybridflow}, Real~\citep{mei2025real}, and RLHFuse~\citep{zhong2025optimizing} place training and rollout on the same GPU pool and alternate between trajectory generation and optimization.
This design guarantees the on-policy training, but it suffers from long-tail rollout latency, leaving expensive trainer GPUs idle during rollout.
\textbf{Disaggregated RL systems} such as AReaL~\citep{fu2025areal} and SLIME~\citep{slime_github} address this utilization problem by decoupling rollout generation from policy optimization, allowing rollout workers and trainers to run on separate GPU pools and overlap execution.
However, the heterogeneity between rollout and training in LLM RL complicates scheduling, resource allocation, and synchronization, while enabling optimizations such as elastic scaling~\citep{wu2025rlboost} and heterogeneous resource management~\citep{yan2025areal, he2025hetrl}.

% TODO: Add related work on multi-agent serving systems, data algorithms for LLM RL,
% weight synchronization / sparse updates, and cluster resource management.

% \input{tex/03_method}
\section{Dataflow-Oriented RL for Agentic LLMs}
\label{sec:dataflow-rl-alt}
In this section, we present the design of \sys.
We begin by motivating the shift from trainer-centered control to dataflow-oriented coordination, explaining why compute decoupling alone is insufficient for agentic RL in Section~\ref{sec:motivation}.
We then introduce the three abstraction designs, as shown in Figure~\ref{fig:astraflow-overview}: a \emph{dataflow layer} that manages rollouts and training batches (Section~\ref{sec:alt-dataflow}); a \emph{Rollout-as-a-Service} abstraction that decouples trajectory generation from optimization (Section~\ref{sec:alt-raas}); and a \emph{trainer} abstraction that consumes batches, updates policies, and publishes weights (Section~\ref{sec:alt-trainer-weight}).
% All components run independently and interact only through minimal data and weight interfaces.
\subsection{Motivation: From Trainer-Centered Control to Dataflow-Oriented Coordination}
\label{sec:motivation}

\textbf{Compute decoupling is not enough.}
Although disaggregated RL frameworks~\citep{fu2025areal, slime_github} separate rollout and training computation, this separation is primarily a compute-placement mechanism, not a principled component abstraction. Rollout scheduling, data selection, replay, staleness handling, and weight synchronization often remain embedded in a trainer-centered control loop. As a result, new capabilities tend to require feature-specific system changes. Multi-policy collaborative training, for instance, requires coordinating multiple independently trained policies, trainers, and weight streams. Elastic, heterogeneous, or cross-region rollouts require additional mechanisms for workers to join and leave dynamically and for weights to be transferred under resource constraints. These capabilities can be added to existing systems, but usually through ad hoc patches or substantial redesign, like Areal-Hex~\citep{yan2025areal}; composing several of them only amplifies the complexity.
\textbf{The root cause behind this limitation is the lack of clean abstraction boundaries among rollout execution, dataflow management, training, and weight transfer.}
Without these boundaries, new capabilities cannot be supported naturally by the system design and instead require explicit feature-specific engineering.
Table~\ref{tab:system_comparison} summarizes this abstraction gap: existing LLM RL frameworks may support asynchrony or rollout-training disaggregation, but generally lack abstractions for composing them.

\textbf{Dataflow-oriented coordination.}
% Motivated by this abstraction gap, we present dataflow-oriented RL for agentic LLMs and its implementation in \sys.
% The key idea is to replace the trainer-centered control loop used by conventional RL systems with dataflow-oriented coordination: rollout services, trainers, and the dataflow layer run \emph{autonomous control loops} and coordinate through minimal data and weight interfaces.
% This abstraction turns rollout execution, training, data management, and weight transfer into composable system boundaries,
% allowing capabilities such as multi-policy collaborative training, elastic rollout pools, heterogeneous and cross-region rollout, and modular data algorithms to be supported natively rather than through feature-specific engineering.
Motivated by this abstraction gap, we propose dataflow-oriented RL for agentic LLMs, a design principle implemented in \sys.
The key insight is that disaggregation should not only separate rollout and training computation, but also separate their control responsibilities.
Instead of organizing the system around a trainer-centered control loop, \sys uses dataflow-oriented coordination: rollout services, trainers, and the dataflow layer each run \emph{autonomous control loops} and interact only through minimal data and weight interfaces.
These interfaces turn rollout, training, data management, and weight transfer into composable system boundaries.
As a result, capabilities such as multi-policy collaborative training, elastic rollout pools, heterogeneous and cross-region rollouts, and modular data algorithms can be expressed by the system architecture itself, rather than added through feature-specific engineering.

\textbf{Design challenges.}
To realize this design, \sys must address three challenges.
First, it must provide a data coordination layer that manages prompts, trajectories, rewards, batching, routing, replay, and staleness across multiple rollout services and trainers without returning control to a single trainer loop.
Second, it must expose stable component boundaries so that rollout engines, trainer backends, and data algorithms can be replaced or extended without pipeline rewrites.
Third, it must support asynchronous and bandwidth-efficient weight flow across multiple policies and rollout pools, including elastic, heterogeneous, and cross-region deployments.

% \begin{figure}[t]
% \centering
% \includegraphics[width=\linewidth]{figs/astra-diagram.pdf}
% \caption{Overview of \sys as a dataflow-oriented RL system.}
% \label{fig:astraflow-overview-alt}
% \end{figure}

\subsection{Dataflow Layer Abstraction}
\label{sec:alt-dataflow}

\begin{wrapfigure}{r}{0.5\textwidth}
\centering
\includegraphics[width=0.5\textwidth]{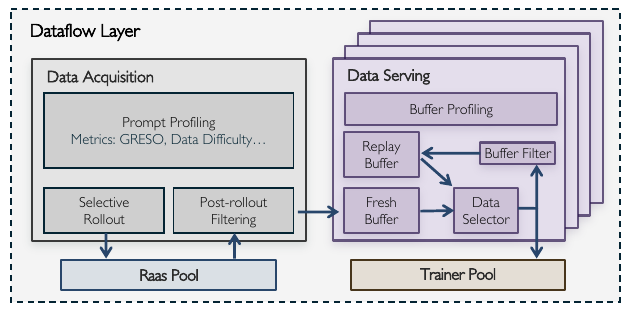}
\caption{Dataflow layer abstraction. Prompt sources, RaaS nodes, and trainers interact through a shared layer that buffers data and applies sampling, filtering, and routing policies.}
\label{fig:alt-dataflow-abstraction}
\end{wrapfigure}

The dataflow layer is the coordination plane between rollout services and trainers.
% Instead of letting the trainer directly drive rollout collection, \sys places rollout-side data acquisition and trainer-side data serving behind a shared dataflow abstraction.
The layer represents RL data in its natural units, including prompts, trajectories, metadata, and training batches.
RaaS nodes pull rollout tasks from the layer and push completed trajectories back, while trainers independently pull batches according to their own optimization loops.

\textbf{Data algorithm interface.}
The dataflow layer exposes a programmable interface for algorithms that operate on prompts, trajectories, rewards, and metadata.
Policies such as selective rollout, curriculum scheduling, post-rollout filtering, dynamic sampling~\citep{yu2025dapo}, replay, data mixing, and staleness correction can therefore be implemented as dataflow policies without modifying the trainer, RaaS implementation, or system orchestration.
% These policies can be implemented as dataflow policies, letting users customize data algorithms 

\textbf{Data-driven coordination.}
The dataflow layer also coordinates autonomous rollout services and trainers through data availability and routing.
Although each component runs its own control loop, the layer can regulate their interaction by deciding which rollout tasks, trajectories, and batches each component receives.
For example, it can throttle slow or stale rollout services by assigning fewer tasks, prioritize fresher trajectories for a trainer, or block unsuitable batches through backpressure.
In multi-policy training, trajectory metadata such as producing policy, model version, timestamp, reward statistics, and task type allows the layer to route policy-specific, shared, or mixed data streams to different trainers without requiring direct trainer-to-trainer coordination.

Together, these two roles make the dataflow layer both a modular data-algorithm interface and a control plane for independent components.
New data policies and coordination strategies can be added in the dataflow layer rather than by rewriting a trainer-centered control loop.

\subsection{Rollout-as-a-Service~(RaaS) Abstraction}
\label{sec:alt-raas}

\begin{wrapfigure}{r}{0.3\textwidth}
\centering
\includegraphics[width=0.3\textwidth]{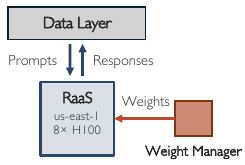}
\caption{RaaS abstraction. Each rollout node consumes tasks, produces trajectories, and refreshes weights.}
\label{fig:alt-raas-abstraction}
\end{wrapfigure}

RaaS models rollout generation as a pure \emph{agent-serving service}.
Each RaaS node receives tasks from the dataflow layer, executes the corresponding agent workflow, and returns trajectories.
% The workflow may be a single LLM response, a multi-turn interaction, a tool-using agent, an environment rollout, or a multi-role procedure such as solver-verifier collaboration.
The RaaS interface only requires that the service consume tasks, produce trajectories, and refresh weights through the trainer-side weight-transfer interface.
This interface makes rollout execution substitutable.
An efficient agent-serving runtime can be plugged into \sys as long as it follows the RaaS contract.
The runtime does not need to know how trajectories are sampled, replayed, filtered, or assigned to trainers.
Likewise, the trainer does not need to know which serving runtime produced the trajectory.
This separation allows \sys to reuse specialized agent-serving systems as rollout backends instead of re-implementing their internal execution logic.

RaaS also makes rollout capacity elastic.
Adding capacity means launching more RaaS nodes that connect to the same dataflow layer and weight-transfer interface.
Removing capacity, slow workers, or failures affect only the rate at which trajectories arrive, not the independent trainer control loop.
This property is especially useful for heterogeneous and cross-region settings, where rollout services may have different latency, throughput, and network bandwidth.

\subsection{Trainer Abstraction and Weight Transfer}
\label{sec:alt-trainer-weight}

\begin{figure}[h]
\centering
\begin{subfigure}[t]{0.3\textwidth}
    \centering
    \includegraphics[width=\linewidth]{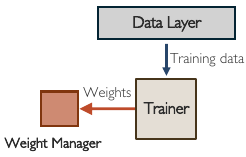}
    \caption{Trainer abstraction.}
    \label{fig:alt-trainer-abstraction}
\end{subfigure}
\hfill
\begin{subfigure}[t]{0.67\textwidth}
    \centering
    \includegraphics[width=\linewidth]{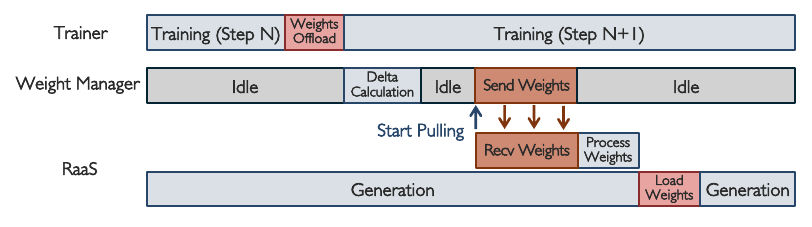}
    \caption{Delta weight transfer.}
    \label{fig:alt-weight-transfer}
\end{subfigure}
\caption{\textbf{(a)} Each trainer consumes batches from the dataflow layer and publishes updated weights through a trainer-side weight interface. \textbf{(b)} Fully async pull-based sparse weight update.}
\label{fig:alt-trainer-and-weight}
\end{figure}

A trainer consumes batches from the dataflow layer, performs optimization with its own backend, and publishes updated weights through a trainer-side weight-transfer interface.
From the trainer's perspective, the dataflow layer behaves like a streaming training corpus, and the weight-transfer interface behaves like a publication target.
The trainer does not need to manage rollout workers, serve model weights directly, or coordinate with other trainers.

This abstraction also makes training backends substitutable.
An existing RL, SFT, or fault-tolerant training backend can participate in \sys if it can consume batches from the dataflow layer and publish weights through the same trainer abstraction.
The same interface also supports multi-policy collaborative training: each policy can have an independent trainer and weight stream, while the dataflow layer controls how trajectories are distributed to each trainer.
% This is different from merely orchestrating multiple rollout roles around one trainable model; the abstraction supports multiple independently optimized policies within the same dataflow.

\textbf{Weight Transfer.} Within the trainer abstraction, the weight-transfer mechanism owns the weight flow between trainers and rollout services.
It stores model versions, exposes the latest or requested versions to RaaS nodes, and handles asynchronous refresh.
Because RaaS nodes pull weights when appropriate, weight delivery is not part of the trainer's critical path.
The trainer-side weight interface can implement full-model transfer, sparse transfer, and version-aware refresh behind the same abstraction.
This design keeps constrained or remote weight transfer isolated from trainer logic while allowing rollout services to refresh at different rates.

% Together, the dataflow layer and trainer-side weight-transfer interface remove the need for a single trainer-centered loop to directly coordinate data movement and model synchronization.
% This is the abstraction gap that \sys addresses: rollout execution, dataflow management, training, and weight transfer become independent services with narrow contracts, so new capabilities can be composed through the system design rather than added as ad hoc patches.

\section{Evaluation: Applications of \sys}
\label{sec:experiments}

In this section, we demonstrate the flexibility of \sys from three perspectives: multi-policy collaborative training, system flexibility, and data algorithm flexibility.
% Our evaluation shows that \sys can support these capabilities within a single unified architecture, without redesigning the training loop or adding ad hoc system extensions for each workload:
% \begin{itemize}
%     \item In Section~\ref{sec:exp-multi-agent}, we evaluate \sys on three two-policy workflows, improving over matched single-policy baselines and achieving a 2.7$\times$ math iteration-time speedup.
%     \item In Section~\ref{sec:exp-autoscale}, we show that rollout auto-scaling can be easily achieved using an agentic maintainer, without requiring any coding.
%     \item In Section~\ref{sec:exp-heter-crossregion}, we show that, \sys can natively support heterogeneous and cross-region training, without feature-specific engineering.
%     \item In Section~\ref{sec:exp-data-flexibility}, we show that, with the data algorithm abstraction in the dataflow layer, we can easily integrate and compose data algorithms like, dynamic sampling, GRESO, and buffer replay.
% \end{itemize}

{\setlength{\leftmargini}{2.25em}
\begin{itemize}
    \item Section~\ref{sec:exp-multi-agent} evaluates \sys on three two-policy workflows, demonstrating improvements over matched single-policy baselines and a 2.7$\times$ speedup in training time.
    \item Section~\ref{sec:exp-autoscale} demonstrates that rollout auto-scaling can be achieved using an agentic maintainer, without requiring any code changes.
    \item Section~\ref{sec:exp-heter-crossregion} highlights \sys's native support for heterogeneous and cross-region training without feature-specific engineering.
    \item Section~\ref{sec:exp-data-flexibility} validates the dataflow-layer abstraction by integrating and composing data algorithms, including dynamic sampling~\citep{yu2025dapo}, GRESO~\citep{zheng2025act}, and buffer replay.
\end{itemize}}
% Across all applications, trainers use GRPO~\citep{shao2024deepseekmath} with FSDP backends and rollout services use SGLang; full hyperparameters, dataset splits, baselines, and per-experiment hardware are deferred to Appendix~\ref{app:experimental_settings}.
% TODO: add SGLang citation once an entry is in references.bib.
Due to the space limitation, we include our detailed experimental setting in Appendix~\ref{app:experimental_settings}.

\subsection{Application I: Multi-Policy Collaborative Training}
\label{sec:exp-multi-agent}

\begin{figure}[h]
\centering
\begin{subfigure}[t]{0.245\textwidth}
    \centering
    \includegraphics[width=0.85\linewidth]{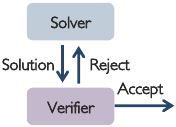}
    \caption{Math: Solver + Verifier.}
    \label{fig:mas-math-verifier}
\end{subfigure}
\hspace{0.035\textwidth}
\begin{subfigure}[t]{0.265\textwidth}
    \centering
    \includegraphics[width=0.85\linewidth]{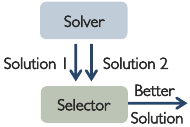}
    \caption{Code: Solver + Selector.}
    \label{fig:mas-code-selector}
\end{subfigure}
% \hspace{0.015\textwidth}
\begin{subfigure}[t]{0.42\textwidth}
    \centering
    \includegraphics[width=0.85\linewidth]{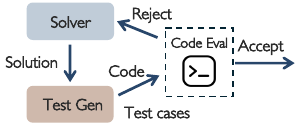}
    \caption{Code: Solver + Test-Case Generator.}
    \label{fig:mas-code-testcase}
\end{subfigure}
\caption{Multi-agent workflows evaluated: \textbf{(a)} Math solver + verifier, \textbf{(b)} Code solver + selector, and \textbf{(c)} Code solver + test-case generator with retry. Different policies serve different roles in a workflow.}
\label{fig:mas-workflows}
\end{figure}

We first evaluate \sys's flexibility on multi-policy collaborative RL through three multi-agent workflows illustrated in Figure~\ref{fig:mas-workflows}: math solver--verifier, code solver--selector, and code solver--test-case generation.
In \sys, users only specify the multi-agent workflow: role order, context passing, and reward assignment.
Existing multi-agent RL systems such as Dr. MAS~\citep{feng2026dr} and Stronger MAS~\citep{zhao2025stronger} build this support on top of verl~\citep{sheng2024hybridflow}, requiring pipeline-level modifications to coordinate multiple roles, policies, trainers, and weight streams, while also inheriting verl's colocated synchronous execution.
In contrast, \sys natively supports multi-policy workflows through its dataflow-oriented RL abstraction.
As a result, multi-agent RL becomes a workflow-level change rather than a system-level pipeline modification.
To the best of our knowledge, \sys is the first LLM RL framework to support fully asynchronous multi-policy collaborative training.

\vspace{0.4cm}
\vspace{0.4cm}

\textbf{Multi-agent Workflow.}
% \ion{What is the target hardware?}
For every multi-policy run, both roles are initialized from Qwen3-8B~\citep{yang2025qwen3} and trained as separate policies.
% The matched single-policy and multi-policy runs share model initialization, rollout budgets, and trainer hyperparameters.
In the math workflow, a Solver proposes a solution and a Verifier accepts or rejects it; if rejected, the Solver receives feedback and retries.
In the code-selector workflow, the Solver generates two candidate programs and the Selector chooses one to submit.
In the code test-case workflow, the Solver writes a program, the Test-Case Generator produces tests, an evaluator executes the program on those tests, and the Solver retries on failure.
Full prompt templates for each role are given in Appendix~\ref{app:multi-agent-prompts}.

\vspace{0.2cm}
\begin{table}[h]
    \centering
    \caption{Math multi-policy training results under matched conditions. Solver and Solver + Verifier (verl) accuracy follows Dr. MAS. \sys achieves comparable or better accuracy while reducing iteration time by 2.7$\times$.}
    \label{tab:multi-agent-math-wods}
    \small
    \setlength{\tabcolsep}{3pt}
    \begin{tabular}{lccccc|c}
    \toprule
    \textbf{Method} & \textbf{AIME24} & \textbf{AIME25} & \textbf{MATH500} & \textbf{Minerva} & \textbf{Avg Acc.} & \textbf{Time/iter (s)} \\
    \midrule
    Solver & 42.9 & 31.8 & 90.5 & 39.2 & 51.1 & --- \\
    Solver + Verifier (verl) & 44.6\,{\scriptsize(+1.7)} & \textbf{41.5}\,{\scriptsize(+9.7)} & 90.7\,{\scriptsize(+0.2)} & 40.9\,{\scriptsize(+1.7)} & 54.4\,{\scriptsize(+3.3)} & 212.64 \\
    Solver + Verifier (\sys) & \textbf{47.3}\,{\scriptsize(+4.4)} & 40.6\,{\scriptsize(+8.8)} & \textbf{92.9}\,{\scriptsize(+2.4)} & \textbf{45.0}\,{\scriptsize(+5.8)} & \textbf{56.5}\,{\scriptsize(+5.4)} & \textbf{77.65} \\
    \bottomrule
    \end{tabular}
\end{table}
\vspace{0.2cm}

% \textbf{\sys speeds up multi-policy math training by 2.7$\times$.}
\textbf{2.7$\times$ speeds up on multi-agent math training.}
Table~\ref{tab:multi-agent-math-wods} reports the math multi-agent training results. Both solver--verifier configurations outperform the single-policy Solver baseline, indicating that collaborative multi-policy training provides an effective learning signal. In particular, \sys improves average accuracy from 51.1\% to 56.5\%, a gain of 5.4\%, which is comparable to or better than the verl-based Dr. MAS implementation. At the same time, \sys is substantially more training-efficient than Dr. MAS: because Dr. MAS inherits verl's colocated synchronous execution, long-tail multi-agent rollouts can stall the entire iteration. By contrast, \sys reduces the time per iteration from 212.64s to 77.65s.

% \begin{table}[h]
% \centering
% \caption{Mathematical-reasoning accuracy of single-policy vs.\ multi-policy collaborative training with Qwen3-8B. The multi-policy run jointly trains an actor and a verifier in one \sys job. Since the AMC split differs for Dr. MAS, we also report an average excluding AMC.}
% \label{tab:multi-agent-math}
% \small
% \begin{tabular}{lccccccc}
% \toprule
% \textbf{Method} & \textbf{AIME24} & \textbf{AIME25} & \textbf{AMC} & \textbf{MATH500} & \textbf{Minerva} & \textbf{Avg} & \shortstack{\textbf{Avg}\\\textbf{w/o AMC}} \\
% \midrule
% Single-policy & 44.2 & 35.2 & 65.8 & 92.4 & 43.1 & 56.1 & 53.7 \\
% Single-policy (w/o DS) & 42.9 & 31.8 & --- & 90.5 & 39.2 & --- & 51.1 \\
% Dr. MAS & 44.6 & 41.5 & 87.5 & 90.7 & 40.9 & 61.0 & 54.4 \\
% Multi-policy (no filter) & 45.0 & 34.6 & 68.2 & 91.4 & 44.5 & 56.7 & 53.9 \\
% Multi-policy (w/o DS) & 47.3 & 40.6 & 68.3 & 92.9 & 45.0 & 58.8 & 56.5 \\
% Multi-policy  & \textbf{50.6} & \textbf{40.6} & \textbf{70.7} & \textbf{93.3} & \textbf{46.9} & \textbf{60.4} & \textbf{57.9} \\
% \bottomrule
% \end{tabular}
% \end{table}

\begin{wrapfigure}{r}{0.33\textwidth}
    \centering
    \includegraphics[width=\linewidth]{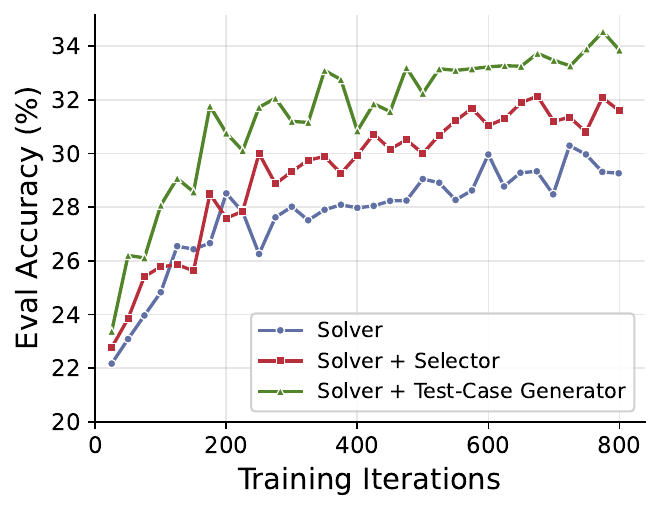}
    \caption{Eval accuracy during training (Qwen3-8B), averaged over LiveCodeBench v5/v6 and Codeforces.}
    \label{fig:code-2agent-eval-curve}
\end{wrapfigure}
\textbf{Generality to code multi-agent workflows.}
Table~\ref{tab:multi-agent-code} evaluates whether the same multi-policy abstraction extends beyond math to code generation.
We consider two interaction patterns: Solver + Selector, where the Solver generates two candidates and the Selector chooses one to submit, and Solver + Test-Case Generator, where generated tests provide execution feedback for retry.
Both workflows improve over the matched single-policy Solver on every benchmark, with the stronger workflow raising average accuracy from 30.29\% to 34.55\% ($+4.26$).
We use the matched Solver baseline rather than a specialized code-agent baseline because this experiment is intended to test system generality under controlled training conditions; Table~\ref{tab:multi-agent-math-wods} provides the direct system-to-system efficiency comparison.
The result shows that \sys can express substantially different multi-agent code workflows by changing only workflow logic, while reusing the same trainer, rollout, dataflow, and weight interfaces.
Figure~\ref{fig:code-2agent-eval-curve} plots the training-time eval accuracy averaged over the three code benchmarks.

\begin{table}[h]
\vspace{0.2cm}
\centering
\caption{Code-generation accuracy of single-policy vs.\ multi-policy collaborative training with Qwen3-8B~\citep{yang2025qwen3}. We report two multi-policy workflows.}
\label{tab:multi-agent-code}
\small
\begin{tabular}{lcccc}
\toprule
\textbf{Method} & \textbf{LiveCodeBench v5} & \textbf{LiveCodeBench v6} & \textbf{Codeforces} & \textbf{Avg} \\
\midrule
Solver     & 36.83 & 32.86 & 21.20 & 30.29 \\
Solver + Selector  & 38.32\,{\scriptsize(+1.49)} & 35.43\,{\scriptsize(+2.57)} & 22.67\,{\scriptsize(+1.47)} & 32.14\,{\scriptsize(+1.85)} \\
Solver + Test-Case Generator  & \textbf{41.62}\,{\scriptsize(+4.79)} & \textbf{36.29}\,{\scriptsize(+3.43)} & \textbf{25.74}\,{\scriptsize(+4.54)} & \textbf{34.55}\,{\scriptsize(+4.26)} \\
\bottomrule
\end{tabular}
% \vspace{0.2cm}
\end{table}

% \begin{wrapfigure}{r}{0.33\textwidth}
% \centering
% \includegraphics[width=\linewidth]{figs/code_2agent_eval_avg.pdf}
% \caption{Eval accuracy during training (Qwen3-8B), averaged over LiveCodeBench v5/v6 and Codeforces.}
% \label{fig:code-2agent-eval-curve}
% \end{wrapfigure}
% Figure~\ref{fig:code-2agent-eval-curve} plots the training-time eval accuracy averaged over the three code benchmarks. All three runs start from the same Qwen3-8B initialization and follow nearly identical curves for the first ${\sim}100$ iterations; the multi-policy runs then separate from the Solver baseline and the gap continues to grow through 800 iterations. This shows that the gains in Table~\ref{tab:multi-agent-code} are not an end-of-training artifact but a sustained training-dynamics effect: collaborative supervision from the Selector or the Test-Case Generator provides a richer learning signal that the Solver alone cannot recover.

\paragraph{Takeaway.}
Across math and code, \sys supports multiple two-policy workflows through the same dataflow-oriented abstraction, improving over matched single-policy baselines while achieving a 2.7$\times$ iteration-time speedup in the direct math comparison.

\subsection{Application II: System Flexibility}
\label{sec:exp-system-flexibility}

We next evaluate \sys's system flexibility through two case studies: (i) zero-code rollout auto-scaling driven by an agentic maintainer (Section~\ref{sec:exp-autoscale}), and (ii) heterogeneous and cross-region training over throttled GPUs and limited-bandwidth links (Section~\ref{sec:exp-heter-crossregion}).
For both case studies, no feature-specific engineering is required: the same RaaS abstraction (Section~\ref{sec:alt-raas}) and weight-transfer runtime (Section~\ref{sec:alt-trainer-weight}) absorb auto-scaling, GPU heterogeneity, and slow remote synchronization without any change to the \sys system.

\subsubsection{Case Study I: Rollout Auto-Scaling with an Agentic Maintainer}
\label{sec:exp-autoscale}

Although \sys does not control the trainer or the rollout services directly, the dataflow layer sits on the path that connects them and can therefore observe the workload balance: how many trajectories each rollout pool produces, how many the trainer consumes, and how long the trainer spends waiting on data.
Every $K$ trainer steps, \sys exports a summary of these three quantities, the trainer waiting fraction $w$, the recent production and consumption counts $n_p$ and $n_c$, and the availability of each rollout service, and uses them to derive a target rollout-pool size with a simple three-zone policy, applying a dead band $[\tau_{\mathrm{low}}, \tau_{\mathrm{high}}]$ on $w$:

\begin{equation}
G_{\text{target}} =
\begin{cases}
\lceil G / (1 - w) \rceil & \text{if } w > \tau_{\mathrm{high}}, \\
\min\!\big(G,\, \lceil G \cdot (n_c/n_p) \cdot \rho \rceil\big) & \text{if } w < \tau_{\mathrm{low}},\; n_p,n_c > 0, \\
G & \text{otherwise},
\end{cases}
\end{equation}
% \vspace{0.2cm}

where $G$ is the current rollout GPU count and $\rho$ is the shrink-margin factor.
The policy adds GPUs when the trainer is starving ($w>\tau_{\mathrm{high}}$), removes GPUs when the pool is over-allocated ($w<\tau_{\mathrm{low}}$).

\vspace{0.2cm}
\begin{table}[h]
    \centering
    \caption{
    Accuracy, training time, and GPU-hour comparison across the three rollout-pool configurations on the Qwen3-14B~\citep{yang2025qwen3} math job.
    Auto-scaling keeps wall-clock time close to the over-allocated baseline while reducing total GPU-hours.
    Bold marks the best GPU-hour costs.}
    \label{tab:autoscale-gpuh}
    \small
    \setlength{\tabcolsep}{4pt}
    \begin{tabular}{lcccccc}
    \toprule
    \textbf{RaaS Strategy} & \textbf{Avg Acc.} & \textbf{Wall (h)} & \textbf{Rollout GPU-h} & \textbf{Trainer GPU-h} & \textbf{Total GPU-h} & \textbf{Wait frac.} \\
    \midrule
    Fixed 6 GPUs    & 68.6 & 35.8 & 214.8 & 143.2 & 358.0 & 26.9\% \\
    Fixed 11 GPUs   & 68.0 & 23.9 & 263.4 & 95.8 & 359.2 &  2.1\% \\
    Auto-scaling    & 67.9 & 24.4 & \textbf{214.5} & 97.4 & \textbf{312.0} &  3.0\% \\
    \bottomrule
    \end{tabular}
\end{table}
\vspace{0.2cm}

\textbf{No-code auto-scaling via an agentic maintainer.}
The RaaS abstraction (Section~\ref{sec:alt-raas}) makes each rollout service hot-swappable: the dataflow layer treats instances as interchangeable producers and tolerates them joining or leaving at runtime, so resizing the rollout pool reduces to launching or retiring RaaS instances on demand.
On top of this, the utilization report above already names a target pool size $G_{\text{target}}$, leaving the maintainer with a purely operational job: read the report and act on the suggested target.
We use Claude Code as the agentic maintainer in our experiment: it runs alongside the training job, polls the utilization summary every $K$ steps, follows the suggested $G_{\text{target}}$, and issues the corresponding cluster commands to launch or retire RaaS instances---preferentially removing unavailable or persistently low-throughput services on scale-down.
\sys contains no scheduler-specific code; porting the loop to Slurm, Kubernetes, or an in-house scheduler only requires adapting the maintainer's instructions to the corresponding cluster interface.

\textbf{Auto-scaling minimizes GPU-hours with comparable accuracy.}
We compare auto-scaling against two fixed rollout-pool baselines that match the lower and upper ends of the controller's scaling range: 6 rollout GPUs and 11 rollout GPUs.
Table~\ref{tab:autoscale-gpuh} reports the resulting accuracy, wall-clock time, trainer wait fraction, and GPU-hour cost.
The 6-GPU baseline uses few rollout resources but starves the trainer, while the 11-GPU baseline removes most trainer waiting by keeping the larger pool allocated for the full run.
Auto-scaling achieves comparable accuracy and nearly the same wall-clock time as the 11-GPU setting, with a small increase from 23.9h to 24.4h.
However, by releasing rollout capacity when it is not needed, it reduces total GPU-hours to 312.0, about 13\% lower than both fixed-pool baselines.
Figure~\ref{fig:mraas-combined} traces the loop in action: the maintainer scales up when trainer waiting rises, holds inside the dead band, and scales down when sustained low waiting indicates over-allocation.

\subsubsection{Case Study II: Heterogeneous and Simulated Cross-Region Training}
\label{sec:exp-heter-crossregion}

\begin{figure}[t]
    \centering
    \begin{subfigure}[t]{0.24\textwidth}
        \centering
        \includegraphics[width=\linewidth]{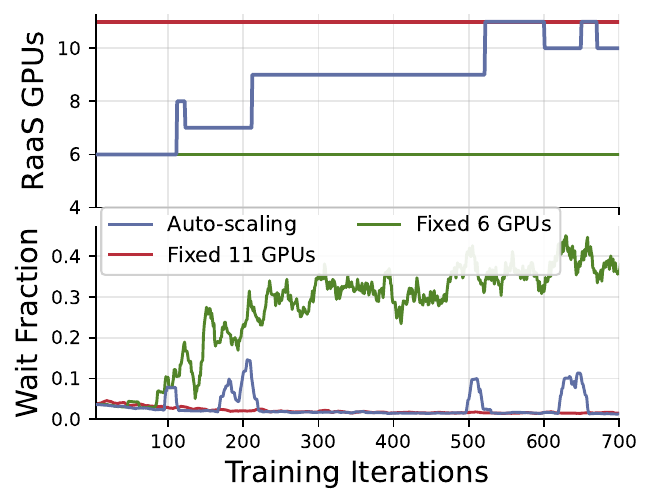}
        \caption{}
        \label{fig:mraas-combined}
    \end{subfigure}
    \hfill
    \begin{subfigure}[t]{0.24\textwidth}
        \centering
        \includegraphics[width=\linewidth]{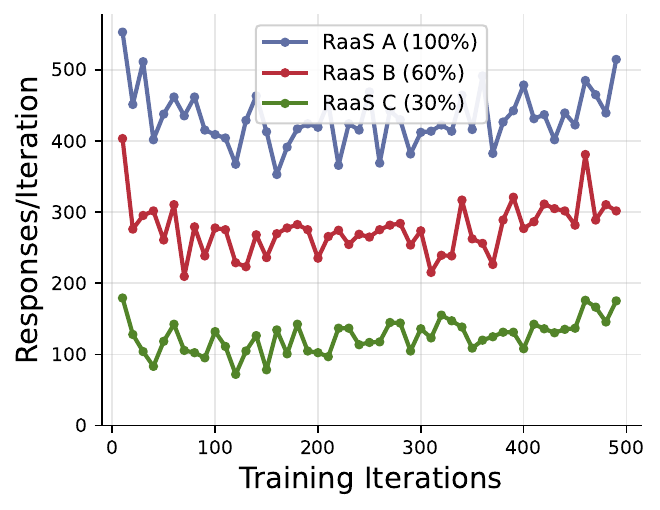}
        \caption{}
        \label{fig:cross-region-produced}
    \end{subfigure}
    \hfill
    \begin{subfigure}[t]{0.24\textwidth}
        \centering
        \includegraphics[width=\linewidth]{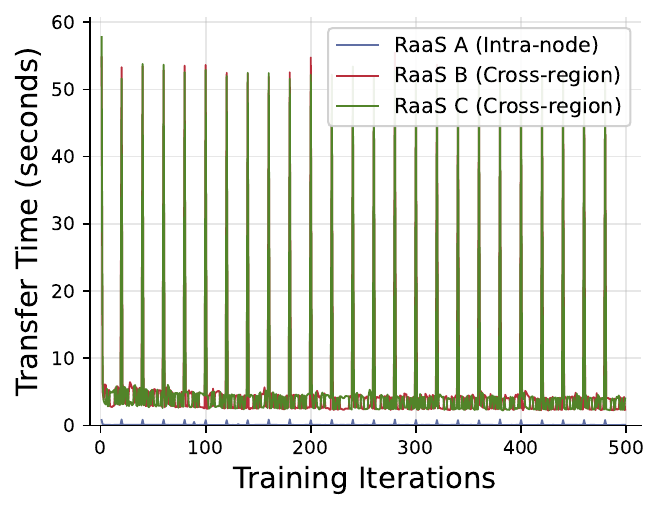}
        \caption{}
        \label{fig:cross-region-transfer}
    \end{subfigure}
    \hfill
    \begin{subfigure}[t]{0.24\textwidth}
        \centering
        \includegraphics[width=\linewidth]{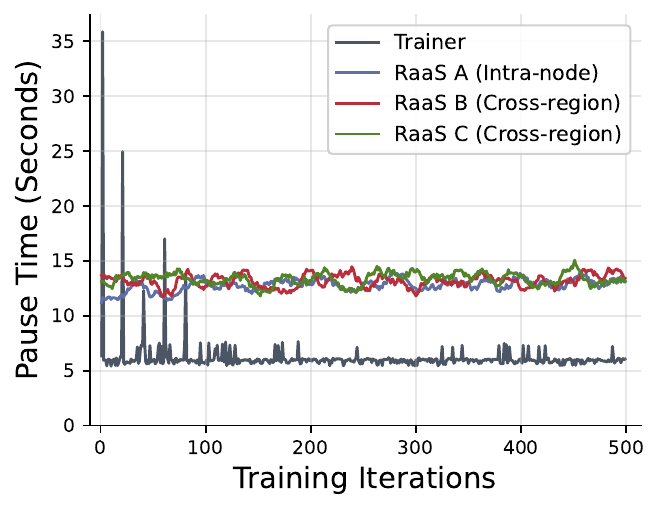}
        \caption{}
        \label{fig:cross-region-downtime}
    \end{subfigure}
    \caption{System-flexibility experiments. \textbf{(a)} Rollout GPU count and trainer waiting time under different rollout-pool settings. \textbf{(b)} Per-iteration rollout throughput from different RaaS pools. \textbf{(c)} Weight-transfer time across local and remote rollout pools. \textbf{(d)} Trainer and rollout downtime in the heterogeneous cross-region setting.}
    \label{fig:system-flexibility}
    \end{figure}

We simulate a cross-region heterogeneous deployment using three nodes (a four-GPU trainer plus three four-GPU RaaS pools, one local and two remote) and train Qwen3-14B~\citep{yang2025qwen3} with M2PO on math.
GPU heterogeneity is induced by per-GPU power caps---700~W on the local pool and 400~W and 250~W on the two remote pools---yielding rollout-throughput shares of roughly 100\%, 60\%, and 30\% respectively in Figure~\ref{fig:cross-region-produced}.
To emulate the cross-region paths, the two remote links are further shaped to 4~Gbit/s effective bandwidth and 300~ms round-trip latency.
All three pools contribute to every iteration and the trainer never blocks on any single pool.

\begin{wrapfigure}{r}{0.33\textwidth}
    \centering
    \includegraphics[width=\linewidth]{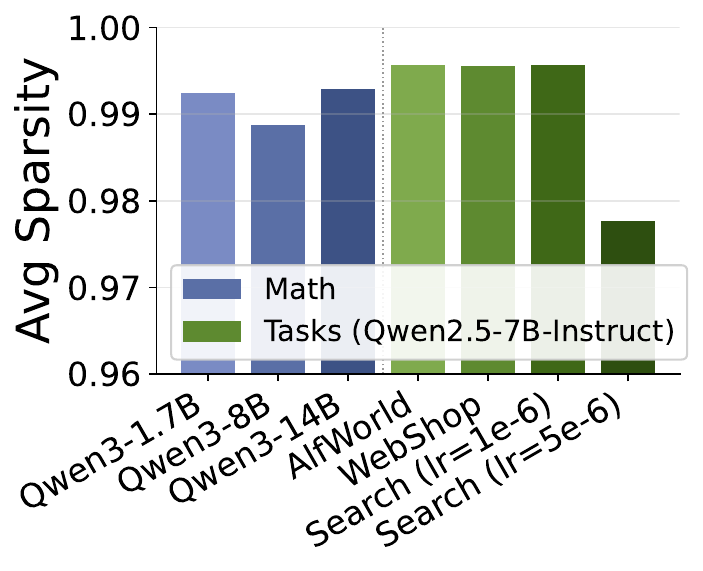}
    \caption{
        % Average bf16-bit-exact delta sparsity (steps {<}500) across model scales (Qwen3-1.7B/8B/14B on math) and tasks (Qwen2.5-7B-Instruct on AlfWorld, WebShop, Search). Higher is more compressible.
    Weight delta sparsity on different models and tasks.
    }
    \label{fig:weight-transfer-sparsity-bar}
\end{wrapfigure}

Training quality is preserved despite the constrained network: the cross-region run reaches 67.6 average accuracy on the five-benchmark math suite, comparable to a homogeneous local baseline.
Figure~\ref{fig:cross-region-transfer} reports per-iteration weight-transfer time on each link.
With $\geq 98.9\%$ delta sparsity for Qwen3-14B~\citep{yang2025qwen3} (Figure~\ref{fig:weight-transfer-sparsity-bar}), the per-iteration payload drops from a $\sim$28~GB full sync to roughly 1.5~GB of delta bytes, so most remote transfers complete in tens of seconds even at 4~Gbit/s and 300~ms RTT.
The periodic full syncs (every 20 iterations) appear as the visible spikes; they remain bounded because they are amortized over many delta steps.

Figure~\ref{fig:cross-region-downtime} shows that even the slow full-sync transfers do not block the training loop.
On the rollout side, downtime is essentially constant across the run: just the SGLang reload and prefill window after each weight update, independent of how the bytes arrived.
On the trainer side, the early iterations show occasional waiting because the previous iteration's full transfer is still finishing on a remote pool; as training progresses and generation lengths grow, the training phase becomes long enough to fully overlap the next iteration's transfer, and trainer downtime converges to a near-constant baseline.
This is the behavior the request-based delta-pull design is intended to deliver (Section~\ref{sec:alt-trainer-weight}): the cost of a slow link is masked by ongoing training work, so heterogeneous rollout and slow remote synchronization are absorbed by the existing abstractions without any change to the trainer, RaaS, or dataflow interfaces.

\textbf{Remote Sparse Weight Transfer.}
As discussed in recent works~\citep{fireworks2026frontierrl, miahi2026understanding}, per-iteration weight updates in RL training are extremely sparse: under bf16 most parameters are bit-exactly identical from one iteration to the next, so the trainer needs to ship only a tiny fraction of its weights to the rollout fleet.
Figure~\ref{fig:weight-transfer-sparsity-bar} measures this directly across two axes: model scale (Qwen3-1.7B / 8B / 14B~\citep{yang2025qwen3} on math) and task (Qwen2.5-7B on AlfWorld, WebShop, and Search), reporting the fraction of bf16 parameters bit-exactly equal to the previous iteration, averaged over the first 500 training iterations.
At a fixed learning rate, sparsity is largely independent of model size and task: every math run lands in $0.989$--$0.993$ and the Qwen2.5-7B tasks all reach $\geq 0.996$ at $\text{lr}=1\mathrm{e}{-6}$.
Learning rate is the dominant driver: raising it to $5\mathrm{e}{-6}$ on Search drops sparsity to $0.978$, since larger updates push more parameters past the bf16 ULP boundary, but even in this most aggressive regime sparsity still exceeds $0.97$, leaving a $\geq 30\times$ compression upper bound on the bytes the trainer must ship per iteration.
The request-based remote weight-transfer path can therefore rely on a high-sparsity assumption across all workloads we measured; per-iteration sparsity curves are deferred to Appendix~\ref{app:delta-sparsity-curves}.

% \paragraph{Takeaway.}
% Without changing a line of \sys code, the same system runs an auto-scaled rollout pool driven by an external controller, deploys across heterogeneous GPUs and regions, and reduces per-step weight-transfer cost by more than an order of magnitude through request-based remote sparse delta transfer.

\subsubsection{Performance Comparison to Existing RL Framework}
\label{sec:exp-perf-vs-existing}

\begin{table}[h]
    \vspace{0.2cm}
    \centering
    \caption{Performance comparison against AReaL on Qwen3-1.7B and Qwen3-8B~\citep{yang2025qwen3} M2PO math reasoning tasks training for 800 iterations. \sys matches AReaL's accuracy and per-iteration efficiency at both scales while providing all the flexibility shown in the previous sections at no cost.
    }
    \label{tab:perf-vs-areal}
    \small
    \setlength{\tabcolsep}{4pt}
    \resizebox{\linewidth}{!}{%
    \begin{tabular}{llcccccc|cc}
    \toprule
    \textbf{Model} & \textbf{Framework} & \textbf{AIME24} & \textbf{AIME25} & \textbf{AMC} & \textbf{MATH500} & \textbf{Minerva} & \textbf{Avg} & \textbf{Time/iter (s)} & \textbf{Time/1M tok (s)} \\
    \midrule
    \multirow{2}{*}{Qwen3-1.7B} & AReaL & 32.5 & 27.5 & 61.1 & 88.2 & 38.2 & 49.5 & 81.5 & 70.1 \\
                                & \sys  & 33.3 & 30.6 & 59.2 & 87.5 & 35.8 & 49.3 & 81.1 & 69.4 \\
    \midrule
    \multirow{2}{*}{Qwen3-8B}   & AReaL & 60.0 & 55.0 & 72.2 & 94.6 & 45.2 & 65.4 & 137.0 & 117.6 \\
                                & \sys  & 62.3 & 50.0 & 72.5 & 94.9 & 44.5 & 64.8 & 139.6 & 119.6 \\
    \bottomrule
    \end{tabular}}
    \vspace{0.2cm}
\end{table}

We train Qwen3-1.7B and Qwen3-8B~\citep{yang2025qwen3} math jobs with M2PO on \sys and on AReaL~\citep{fu2025areal}, a representative trainer-centric RL framework, and compare accuracy and training speed (Table~\ref{tab:perf-vs-areal}).
At both scales, the two systems reach comparable accuracy on the math suite (within 0.2 / 0.6 points on average) and comparable per-iteration training speed (within 1\% / 2\%): \sys matches AReaL's performance and efficiency on a workload AReaL is specialized for, while still providing the flexibility demonstrated in the previous sections.

\subsection{Application III: Data Algorithm Flexibility}
\label{sec:exp-data-flexibility}

We train Qwen3-8B~\citep{yang2025qwen3} on math with three representative data algorithms, chosen to cover the three intervention points along the RL data path: \emph{GRESO}~\citep{zheng2025act} performs \textbf{selective rollout}, deciding which prompts are sent to generation; \emph{dynamic sampling}~\citep{yu2025dapo} performs \textbf{post-rollout filtering}, discarding zero-advantage trajectories after generation; and \emph{buffer replay} performs \textbf{training-batch selection}, reusing useful trajectories across multiple trainer updates.
Together they exercise pre-, post-, and serving-side hooks of the dataflow layer.

\begin{wrapfigure}{r}{0.33\textwidth}
    \centering
    \includegraphics[width=\linewidth]{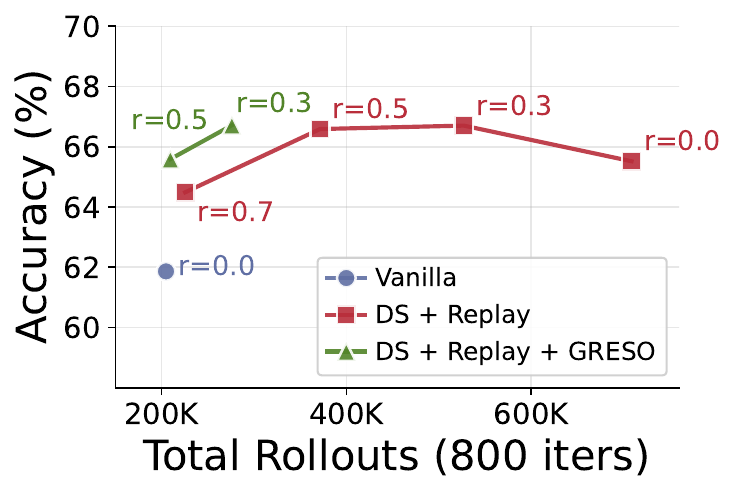}
    \caption{Math accuracy versus generated rollouts for dataflow-layer data algorithms.}
    \label{fig:math-buffer-greso-acc-rollouts}
\end{wrapfigure}
Figure~\ref{fig:math-buffer-greso-acc-rollouts} reports accuracy versus generated rollouts for the three algorithms.
Dynamic sampling~\citep{yu2025dapo} lifts final accuracy meaningfully but increases generation cost by roughly 3.5$\times$ (about 200k $\rightarrow$ 700k rollouts), since post-filtering discards a large fraction of generations.
GRESO~\citep{zheng2025act} and buffer replay sit on the opposite side of the trade-off: both reach baseline-level accuracy with substantially fewer generated rollouts---GRESO~\citep{zheng2025act} by avoiding low-value prompts before rollout, buffer replay by reusing trajectories at batch serving.
This data algorithm composition shows that: each algorithm is a self-contained policy class that the dataflow layer imports as a plug-in, so prompt selection, rollout filtering, and trajectory replay become composable, modular components rather than system-wide rewrites.

\section{Conclusion}
\label{sec:conclusion}

We presented \sys, a dataflow-oriented RL training system for agentic LLMs.
Our central observation is that the rigidity of existing RL training systems comes from a single architectural choice: the trainer sits at the center of the loop and orchestrates everything else.
\sys replaces this trainer-centered control with three narrow abstractions, a dataflow layer for coordination, a rollout service for trajectory generation, and a trainer for policy updates, that interact only through stable interfaces.
Because no component owns the global control flow, properties such as full asynchrony, elastic rollout scaling, plug-in trainers and inference engines, multi-trainer multi-policy coordination, and efficient delta and request-based sparse weight transfer all fall out of the same design rather than being bolted on as separate mechanisms.

% TODO: 1--2 sentences summarizing the headline experimental results once available
% (e.g., end-to-end speedup vs.\ baseline, weight-transfer compression ratio,
% multi-policy collaborative training accuracy, elastic-scaling behavior).

% We hope \sys's abstractions will serve as a stable foundation for future work on agentic RL, including larger multi-agent systems, new data and trust-region algorithms, and deployment across heterogeneous and cross-region compute.
% TODO: optionally add one sentence on broader impact or release plans (open-source code, models, benchmarks).

\bibliographystyle{plainnat}
\bibliography{references}

\clearpage
\beginappendix

\section*{Limitations}
\label{app:limitations}

\sys focuses on system abstractions for agentic RL workloads rather than proposing a new RL optimization algorithm. Our evaluation covers representative math, coding, multi-policy, elastic rollout, heterogeneous deployment, and data-algorithm workloads, but does not exhaust all agentic RL settings, such as long-horizon web agents, robotics environments, or safety-critical interactive systems. The observed benefits may depend on workload characteristics including rollout latency, trainer throughput, network bandwidth, weight-transfer frequency, and asynchronous data availability. In addition, some experiments use controlled or simulated deployment settings, and \sys assumes that rollout tasks, trajectory metadata, and reward signals can be represented through the dataflow layer. While \sys makes multi-policy and data-centric RL workloads easier to compose, model quality, robustness, and safety still depend on the underlying data, rewards, policies, and evaluation protocol.

\section*{Social Impact}
\label{app:social-impact}

\sys is an infrastructure system for reinforcement learning of agentic language models. Its positive impact is to make advanced RL experimentation more modular and resource-efficient by allowing researchers to reuse rollout services and trainers, compose data algorithms, and run flexible multi-policy or heterogeneous deployments without rewriting the full training pipeline. At the same time, more efficient RL infrastructure may accelerate stronger agentic models for coding, tool use, and autonomous decision-making, which could be misused without appropriate evaluation, monitoring, access control, or domain-specific safeguards. \sys does not release a new foundation model or dataset, but users should still evaluate trained agents for reliability, security, privacy, and misuse risks before deployment.

% \section{Appendix Overview}
% \label{app:overview}

\section{Experimental Settings}
\label{app:experimental_settings}

This appendix provides the full experimental setup deferred from Section~\ref{sec:experiments}.
We organize the appendix per-experiment so that each subsection is self-contained and reproducible.
Section~\ref{app:exp-common} fixes the shared bits that recur across experiments (model cards, evaluation suites, decoding, reward functions);
Sections~\ref{app:exp-multi-agent-math}--\ref{app:exp-data-flexibility} then specify, one per main-text experiment, the training data, algorithm hyperparameters, hardware and topology, baselines, and any experiment-specific knobs.

\subsection{Common Setup}
\label{app:exp-common}

\textbf{Evaluation protocol.}
Across \emph{all} experiments in this appendix, we evaluate each benchmark by sampling 4 independent generations per question (temperature $0.6$, $n_{\mathrm{samples}}=1$) and reporting the mean per-question accuracy averaged over the benchmark.
For every run we evaluate at a fixed cadence during training and report numbers from the checkpoint with the best benchmark-average accuracy.
Per-experiment subsections below therefore omit this protocol and only note any deviations.

% TODO: model cards (Qwen3-1.7B/8B/14B, Qwen2.5-7B-Instruct):
%       parameter counts, tokenizer, context length, HF revision IDs.
% TODO: evaluation suites:
%       - Math: AIME24 (HuggingFaceH4/aime_2024), AIME25, AMC24, MATH500, Minerva.
%       - Code: LiveCodeBench v5, LiveCodeBench v6, Codeforces.
%       - Agentic: AlfWorld, WebShop, search QA (ASearcher).
% TODO: decoding defaults (training vs.\ eval) and reward functions
%       (math_verify rule-based; code exec-based pass/fail).

\subsection{Multi-policy Math (Section~\ref{sec:exp-multi-agent}, Table~\ref{tab:multi-agent-math-wods})}
\label{app:exp-multi-agent-math}

\textbf{Models \& Datasets.}
Both Solver and Verifier are initialized from Qwen3-8B~\citep{yang2025qwen3} and trained as separate policies under the actor-and-verify workflow described in Section~\ref{sec:exp-multi-agent}.
We train on the DAPO RL math set~\citep{yu2025dapo} (Hugging Face: \texttt{aaabiao/dapo\_filter}), filtering prompts to at most $2{,}000$ tokens; following Dr. MAS~\citep{feng2026dr}, the trailing ``Let's think step by step \ldots \textbackslash boxed\{\}'' suffix is stripped at load time so the multi-agent workflow owns response formatting.
Verbatim prompts are listed in Appendix~\ref{app:multi-agent-prompts}.

\textbf{Training.}
Both policies are trained with M2PO ($m^2$-threshold $0.01$), with group-level reward normalization, batch-level advantage normalization, and a fixed KL penalty coefficient of $10^{-3}$ (no KL controller).
We use AdamW with a constant learning rate of $5\mathrm{e}{-6}$ ($\beta_1{=}0.9$, $\beta_2{=}0.999$, weight decay $0.01$), no warmup, and gradient clipping at $1.0$.
The training batch size is $256$ with $8$ rollouts per prompt and $4$ PPO mini-batches per iteration; rollout generation uses temperature $1.0$ and $\max_{\mathrm{new\_tokens}}=4{,}096$.
We deliberately disable dynamic sampling in this run to align with the Dr. MAS~\citep{feng2026dr} no-DS configuration that the Solver and Solver+Verifier (verl) baselines in Table~\ref{tab:multi-agent-math-wods} report against; this keeps the rollout regime matched across the three rows.
Training runs for $1{,}200$ iterations on a single 8$\times$NVIDIA H100 (80\,GB) node, with $4$ GPUs serving the two SGLang RaaS instances (Solver and Verifier each at data-parallel size $2$) and $4$ GPUs split into two FSDP trainer groups of size $2$ (one per policy).
Trainer$\leftrightarrow$RaaS weight transfer runs in TCP delta mode with a full-sync interval of $10$ iterations.

\textbf{Baseline and time measurement.}
The Solver and Solver+Verifier (verl) accuracy rows in Table~\ref{tab:multi-agent-math-wods} are the corresponding numbers reported in Dr. MAS~\citep{feng2026dr} (Qwen3-8B, GRPO over the same DAPO math set, no DS), which we adopt unchanged.
The per-iteration training-time entries ($212.64$\,s for verl, $77.65$\,s for \sys), in contrast, are both measured on the same 8$\times$H100 node used for the \sys run: we re-run the verl-based Dr. MAS pipeline end-to-end on this hardware and time each iteration with eval steps excluded.
For verl the iteration is rollout$+$train run sequentially under colocated synchronous execution; for \sys the iteration is the trainer step time, with rollout overlap absorbed by the dataflow layer.

\textbf{Evaluation.}
We evaluate on AIME24, AIME25, MATH500, and Minerva Math every $25$ training iterations.
Following the shared protocol in Section~\ref{app:exp-common}, we sample $4$ generations per question at temperature $0.6$ and report $\mathrm{pass}@1\,(\mathrm{avg}@4)$.

\subsection{Multi-policy Code (Section~\ref{sec:exp-multi-agent}, Table~\ref{tab:multi-agent-code})}
\label{app:exp-multi-agent-code}

\textbf{Models \& Datasets.}
The three rows of Table~\ref{tab:multi-agent-code} (single-policy Solver, Solver + Selector, Solver + Test-Case Generator) all initialize from Qwen3-8B~\citep{yang2025qwen3} and use the workflows described in Section~\ref{sec:exp-multi-agent}; multi-agent runs train each role as a separate policy.
We train on the DeepCoder-Preview-Dataset~\citep{deepcoder2025} (Hugging Face: \texttt{agentica-org/DeepCoder-Preview-Dataset}, \texttt{primeintellect} subset, \texttt{train} split) and filter prompts to at most $6{,}000$ tokens.
Reward in all three runs is the binary outcome of executing the final attempt against the problem's hidden test cases.
Verbatim Solver, Selector, and Test-Case Generator prompts are listed in Appendix~\ref{app:multi-agent-prompts}.

\textbf{Training.}
All three runs use the same M2PO settings as Section~\ref{app:exp-multi-agent-math}, except: learning rate $3\mathrm{e}{-6}$, $1{,}000$ training iterations, dynamic sampling on (zero-advantage groups are filtered before training), and TCP \emph{full}-sync weight transfer instead of delta.
The training batch size is $256$ with $8$ rollouts per prompt; in the Solver + Selector run the Selector trainer uses a smaller batch size of $128$, matching its lower per-iteration production rate.
All runs use NVIDIA H200 (141\,GB) GPUs.
The Solver baseline runs on a single 8$\times$H200 node, with $4$ GPUs serving the policy (data-parallel size $4$) and $4$ GPUs as one FSDP trainer group of size $4$.
The two multi-agent runs each use two 8$\times$H200 nodes ($16$ GPUs total): one node fully dedicated to RaaS hosting both policies, and the other node split into $4$ RaaS GPUs and $4$ FSDP trainer GPUs (two groups of size $2$, one per policy).
Rollout data-parallel sizes are $6$ + $6$ (Solver + Selector) and $9$ + $3$ (Solver + Test-Case Generator); the asymmetric v3 split reflects the Test-Case Generator's lower per-rollout invocation rate.

\textbf{Evaluation.}
We evaluate on LiveCodeBench v5~\citep{jain2024livecodebench}, LiveCodeBench v6, and the Codeforces split of DeepCoder-Preview-Dataset every $25$ training iterations; multi-agent runs are scored end-to-end through the same workflow used in training.
Following the shared protocol in Section~\ref{app:exp-common}, we sample $4$ generations per question at temperature $0.6$ and report $\mathrm{pass}@1\,(\mathrm{avg}@4)$.

\subsection{Rollout Auto-scaling (Section~\ref{sec:exp-autoscale}, Table~\ref{tab:autoscale-gpuh})}
\label{app:exp-autoscale}

\textbf{Models \& Datasets.}
A single Qwen3-14B~\citep{yang2025qwen3} policy is trained on the DeepScaler RL math set~\citep{deepscaler2025}, filtering prompts to at most $2{,}000$ tokens.

\textbf{Training.}
We use M2PO with $m^2$-threshold $0.002$, group-level reward and batch-level advantage normalization, and a fixed KL penalty coefficient of $10^{-3}$ (no KL controller).
We optimize with AdamW at a constant learning rate of $3\mathrm{e}{-6}$ ($\beta_1{=}0.9$, $\beta_2{=}0.999$, weight decay $0.01$), no warmup, and gradient clipping at $1.0$.
The training batch size is $256$ with $8$ rollouts per prompt and $4$ PPO mini-batches per iteration; rollout generation uses temperature $1.0$ with $\max_{\mathrm{new\_tokens}}=18{,}000$ and SGLang context length $20{,}480$; dynamic sampling is on (zero-advantage groups are filtered before training).
Training runs for $1{,}200$ iterations on two 8$\times$NVIDIA H200 (141\,GB) nodes ($16$ GPUs total).
The trainer is fixed at $4$ GPUs as a single FSDP group of size $4$ on the main node; the remaining $4$ GPUs on the main node and all $8$ GPUs on the secondary node form the rollout pool.
Each SGLang RaaS instance runs at tensor-parallel size $1$ and is sized at data-parallel $1$, $2$, or $4$ so that the controller can compose pool sizes by adding or retiring instances.
Trainer$\leftrightarrow$RaaS weight transfer uses TCP full-sync mode.

\textbf{Auto-scaling controller.}
We compare three rollout-pool configurations: a fixed pool of $6$ GPUs, a fixed pool of $11$ GPUs, and our auto-scaled pool, which the controller resizes between $6$ and $11$ GPUs.
The dataflow layer exports the trainer waiting fraction $w$, recent production/consumption counts, and per-instance availability every $K=10$ trainer steps.
We use the three-zone policy from Section~\ref{sec:exp-autoscale} with lower and upper waiting-fraction thresholds $\tau_{\mathrm{low}}=0.05$ and $\tau_{\mathrm{high}}=0.10$ and a shrink-margin factor of $\rho=1.10$.
The auto-scaling loop is executed by Claude Code as an agentic maintainer: at each report it reads the suggested $G_{\mathrm{target}}$ and issues the corresponding shell commands to launch or retire RaaS instances, preferentially removing unavailable or persistently low-throughput services on scale-down.

\textbf{Balance report.}
The dataflow layer emits the balance report shown below every $K{=}10$ trainer versions. It is the sole signal the agentic maintainer consumes for scaling decisions. The report has three blocks: (i) a \emph{window} block summarizing wall-clock time and the trainer waiting fraction $w=\sum_k \text{wait}_k / \sum_k \text{step}_k$ over the window; (ii) a \emph{production} block aggregating produced/accepted/consumed tokens across all RaaS instances; (iii) a \emph{scaling decision} block reporting the three-zone branch and the target GPU count $G_{\mathrm{target}}$. A per-RaaS layout table follows so the maintainer can preferentially retire suspect or low-throughput instances on scale-down. \texttt{stale\_skipped} is advisory and does not enter the scaling math.

\begin{tcolorbox}[myreport={Balance Report Template}, fontupper=\fontencoding{T1}\ttfamily\footnotesize, breakable]
\begin{Verbatim}[fontsize=\footnotesize]
--- Window (last <N> iterations) ---
wall_time_sec          : <wall_time>
eval_time_sec          : <eval_time>
training_time_sec      : <training_time>
avg_step_time_sec      : <avg_step_time>
avg_batch_wait_sec     : <avg_wait>
rollout_wait_fraction  : <wait_fraction>

--- Production ---
total_raas_gpus        : <G>
produced               : <produced>
entered                : <accepted>
  accept_rate          : <accepted/produced>
consumed               : <consumed>
stale_skipped          : <stale>
  stale_rate           : <stale/accepted>
throughput_per_gpu     : <accepted/G>
produce_consume_ratio  : <accepted/consumed>

--- Scaling decision ---
branch                 : <scale_up | scale_down | hold>
G_target               : <G_target>
estimated_delta_gpus   : <G_target - G, signed>
weight_transfer_active : <true | false>

--- RaaS Instance Layout (last <N> iterations) ---
uid               gpus    produced    accepted   accept_rate  throughput/gpu    status
<uid_1>            <g_1>      <p_1>       <a_1>       <ar_1>         <tpg_1>   healthy
<uid_2>            <g_2>      <p_2>       <a_2>       <ar_2>         <tpg_2>   suspect
...
---
Total: <M> instances, <sum_g> GPUs, <sum_p> produced, <sum_a> accepted
\end{Verbatim}
\end{tcolorbox}

\noindent The three-zone decision rule the agent reads off the \texttt{branch} field is: scale up via $G_{\mathrm{target}}=\lceil G/(1-w)\rceil$ when $w>\tau_{\mathrm{high}}{=}0.10$; scale down via $G_{\mathrm{target}}=\min(G,\lceil G\cdot(\text{consumed}/\text{accepted})\cdot\rho\rceil)$ with $\rho{=}1.10$ when $w<\tau_{\mathrm{low}}{=}0.05$ and the window saw both production and consumption; otherwise hold.

\textbf{Evaluation.}
We evaluate on AIME24, AIME25, AMC, MATH500, and Minerva Math every $50$ training iterations.
Following the shared protocol in Section~\ref{app:exp-common}, we sample $4$ generations per question at temperature $0.6$ and report $\mathrm{pass}@1\,(\mathrm{avg}@4)$.

\subsection{Heterogeneous and Cross-region Training (Section~\ref{sec:exp-heter-crossregion})}
\label{app:exp-heter-crossregion}

\textbf{Models \& Datasets.}
A single Qwen3-14B~\citep{yang2025qwen3} policy is trained on the DeepScaler RL math set~\citep{deepscaler2025}, filtering prompts to at most $2{,}000$ tokens.

\textbf{Training.}
We use M2PO with $m^2$-threshold $0.002$, group-level reward and batch-level advantage normalization, and a fixed KL penalty coefficient of $10^{-3}$ (no KL controller).
We optimize with AdamW at a constant learning rate of $3\mathrm{e}{-6}$ ($\beta_1{=}0.9$, $\beta_2{=}0.999$, weight decay $0.01$), no warmup, and gradient clipping at $1.0$.
The training batch size is $256$ with $8$ rollouts per prompt and $4$ PPO mini-batches per iteration; rollout generation uses temperature $1.0$ with $\max_{\mathrm{new\_tokens}}=18{,}000$; dynamic sampling is on (zero-advantage groups are filtered before training).
Training runs for $1{,}200$ iterations.
All training settings (model, dataset, M2PO hyperparameters, batch, iterations, decoding, dynamic sampling) are identical to the auto-scaling experiment in Section~\ref{app:exp-autoscale}; only the deployment topology (cross-region heterogeneous vs.\ homogeneous local) and the weight-transfer mode (delta vs.\ full sync) differ.
The Fixed-11-GPU row in Table~\ref{tab:autoscale-gpuh} (avg.~accuracy $68.0$) therefore serves as the homogeneous local baseline against which we compare the cross-region run's $67.6$ in Section~\ref{sec:exp-heter-crossregion}.

\textbf{Cross-region topology and heterogeneity.}
We simulate a cross-region heterogeneous deployment over three NVIDIA H200 (141\,GB) nodes: a 4-GPU FSDP trainer (data-parallel size $4$) on the local node, and three 4-GPU SGLang RaaS pools (each at tensor-parallel size $1$, data-parallel size $4$) --- one co-located with the trainer and two on remote nodes.
GPU heterogeneity is induced by per-GPU power caps: $700$\,W on the local pool, and $400$\,W and $250$\,W on the two remote pools, yielding rollout-throughput shares of roughly $100\%$, $60\%$, and $30\%$ respectively (Figure~\ref{fig:cross-region-produced}).
The two remote links are shaped with \texttt{tc/netem} to $4$\,Gbit/s effective bandwidth and $300$\,ms round-trip latency; the local link is unshaped.
Trainer$\leftrightarrow$RaaS weight transfer uses TCP delta mode with a full-sync interval of $20$ iterations; the periodic full syncs surface as the spikes in Figure~\ref{fig:cross-region-transfer}.

\textbf{Evaluation.}
We evaluate on AIME24, AIME25, AMC, MATH500, and Minerva Math every $50$ training iterations.
Following the shared protocol in Section~\ref{app:exp-common}, we sample $4$ generations per question at temperature $0.6$ and report $\mathrm{pass}@1\,(\mathrm{avg}@4)$.

\subsection{Delta-sparsity Measurement (Section~\ref{sec:exp-heter-crossregion}, Figure~\ref{fig:weight-transfer-sparsity-bar})}
\label{app:exp-delta-sparsity}

\textbf{Models \& Datasets.}
We instrument six independent RL training runs that span two axes:
(i) model scale, training Qwen3-1.7B / 8B / 14B~\citep{yang2025qwen3} on the DeepScaler math set~\citep{deepscaler2025};
(ii) task, training Qwen2.5-7B-Instruct~\citep{yang2024qwen2} on AlfWorld, WebShop, and search-based QA via the ASearcher workflow.
All runs filter prompts to at most $2{,}000$ tokens (math) or $4{,}096$ tokens (agentic).

\textbf{Training.}
Each underlying run is trained with M2PO with the same shared base settings as Section~\ref{app:exp-multi-agent-math} (group reward / batch advantage normalization, fixed KL penalty $10^{-3}$, AdamW with $\beta_1{=}0.9$, $\beta_2{=}0.999$, weight decay $0.01$, constant LR with no warmup, gradient clipping $1.0$, $4$ PPO mini-batches, dynamic sampling on, $8$ rollouts per prompt at temperature $1.0$); the per-run differences are summarized in Table~\ref{tab:sparsity-runs}.
For the lr-ablation on Search, we additionally train the same workload at $\mathrm{lr}=5\mathrm{e}{-6}$ to study the effect of larger updates on sparsity.
Hardware is NVIDIA H100 (80\,GB), with a 4-GPU FSDP trainer and a 4-GPU SGLang RaaS pool per run.

\begin{table}[h]
\centering
\caption{Per-run hyperparameters for the delta-sparsity measurement runs.}
\label{tab:sparsity-runs}
\small
\setlength{\tabcolsep}{4pt}
\begin{tabular}{llccccc}
\toprule
\textbf{Workload} & \textbf{Model} & \textbf{lr} & $\boldsymbol{m^2}$ \textbf{thresh.} & \textbf{Batch} & \textbf{Iters} & $\boldsymbol{\max}_{\mathbf{new\_tok}}$ \\
\midrule
DeepScaler math & Qwen3-1.7B            & $5\mathrm{e}{-6}$ & $0.01$  & $256$ & $800$  & $14{,}000$ \\
DeepScaler math & Qwen3-8B              & $5\mathrm{e}{-6}$ & $0.01$  & $256$ & $800$  & $14{,}000$ \\
DeepScaler math & Qwen3-14B             & $3\mathrm{e}{-6}$ & $0.002$ & $256$ & $1{,}200$ & $14{,}000$ \\
AlfWorld        & Qwen2.5-7B-Instruct   & $1\mathrm{e}{-6}$ & $0.004$ & $128$ & $1{,}200$ & $512$ \\
WebShop         & Qwen2.5-7B-Instruct   & $1\mathrm{e}{-6}$ & $0.004$ & $256$ & $1{,}200$ & $512$ \\
Search (ASearcher) & Qwen2.5-7B-Instruct & $1\mathrm{e}{-6}$ & $0.004$ & $256$ & $1{,}000$ & $1{,}024$ \\
\bottomrule
\end{tabular}
\end{table}

\textbf{Measurement.}
For each run we snapshot the trainer's bf16 weights at every iteration and compute, for each consecutive pair of snapshots, the fraction of parameters that are bit-exactly equal to the previous iteration.
Figure~\ref{fig:weight-transfer-sparsity-bar} reports this fraction averaged over the first $500$ training iterations, one bar per run.
Per-iteration sparsity curves over the full training horizon are in Appendix~\ref{app:delta-sparsity-curves}.

\subsection{Performance vs.\ AReaL (Section~\ref{sec:exp-perf-vs-existing}, Table~\ref{tab:perf-vs-areal})}
\label{app:exp-perf-vs-areal}

\textbf{Models \& Datasets.}
We compare \sys against AReaL~\citep{fu2025areal} on two math jobs at different model scales: a single Qwen3-1.7B policy and a single Qwen3-8B policy, both trained on the DeepScaler RL math set~\citep{deepscaler2025} with prompts filtered to at most $2{,}000$ tokens.
Both frameworks use the same model initialization, training data, training budget, and algorithm hyperparameters; only the framework implementation differs.

\textbf{Training.}
Both jobs train with M2PO using the same configuration: $m^2$-threshold $0.01$, group-level reward and batch-level advantage normalization, fixed KL penalty coefficient of $10^{-3}$ (no KL controller), AdamW with a constant learning rate of $5\mathrm{e}{-6}$ ($\beta_1{=}0.9$, $\beta_2{=}0.999$, weight decay $0.01$), no warmup, and gradient clipping at $1.0$.
The training batch size is $256$ with $8$ rollouts per prompt and $4$ PPO mini-batches per iteration; rollout generation uses temperature $1.0$ with $\max_{\mathrm{new\_tokens}}=14{,}000$ and SGLang context length $16{,}384$; dynamic sampling is on (zero-advantage groups are filtered before training).
Each job runs for $800$ iterations on a single 8$\times$NVIDIA H100 (80\,GB) node, with $4$ GPUs serving the policy (data-parallel size $4$, tensor-parallel size $1$) and $4$ GPUs as one FSDP trainer group of size $4$.
Trainer$\leftrightarrow$RaaS weight transfer uses TCP full-sync mode.
The AReaL baseline reuses the exact same configuration --- model initialization, training data, training budget, M2PO hyperparameters, generation settings, hardware layout, and weight-transfer mode --- on the same single 8$\times$H100 node; only the framework implementation differs.

\textbf{Evaluation.}
We evaluate on AIME24, AIME25, AMC, MATH500, and Minerva Math.
Following the shared protocol in Section~\ref{app:exp-common}, we sample $4$ generations per question at temperature $0.6$ and report $\mathrm{pass}@1\,(\mathrm{avg}@4)$.
We additionally report two efficiency numbers: median per-iteration training time, and total wall time normalized by the number of training tokens (Time/1M tok), both averaged after the warmup phase.

\subsection{Data-algorithm Flexibility (Section~\ref{sec:exp-data-flexibility}, Figure~\ref{fig:math-buffer-greso-acc-rollouts})}
\label{app:exp-data-flexibility}

\textbf{Models \& Datasets.}
A single Qwen3-8B~\citep{yang2025qwen3} policy is trained on the DeepScaler RL math set~\citep{deepscaler2025}, capped at the first $8{,}000$ prompts and filtered to at most $2{,}000$ tokens.
We truncate the training set to $8{,}000$ prompts (rather than the full split used in Sections~\ref{app:exp-autoscale}--\ref{app:exp-perf-vs-areal}) because GRESO requires multiple passes over the same prompts to estimate per-prompt zero-variance probabilities and converge its bucket-level submit-probability schedule; capping the set ensures every variant runs through the prompt stream multiple times within the $800$-iteration budget.
This is also why the $y$-axis in Figure~\ref{fig:math-buffer-greso-acc-rollouts} sits a few points below the same-model accuracy reported on the full DeepScaler set in Section~\ref{app:exp-perf-vs-areal}.
All four variants share this identical underlying prompt stream so that any difference in accuracy or rollout cost reflects the data algorithm rather than the data.

\textbf{Training (shared).}
All variants train with M2PO using the same configuration: $m^2$-threshold $0.01$, group-level reward and batch-level advantage normalization, fixed KL penalty coefficient of $10^{-3}$ (no KL controller), AdamW with a constant learning rate of $5\mathrm{e}{-6}$ ($\beta_1{=}0.9$, $\beta_2{=}0.999$, weight decay $0.01$), no warmup, and gradient clipping at $1.0$.
The training batch size is $256$ with $8$ rollouts per prompt and $4$ PPO mini-batches per iteration; rollout generation uses temperature $1.0$ with $\max_{\mathrm{new\_tokens}}=14{,}000$ and SGLang context length $16{,}384$.
Each variant runs for $800$ iterations on a single 8$\times$NVIDIA H100 (80\,GB) node, with $4$ GPUs serving the policy (data-parallel size $4$, tensor-parallel size $1$) and $4$ GPUs as one FSDP trainer group of size $4$; weight transfer uses TCP full-sync mode.

\textbf{Data-algorithm variants.}
Figure~\ref{fig:math-buffer-greso-acc-rollouts} compares three method families that compose dataflow-layer hooks at different intervention points:
(i) \emph{Vanilla} (1 point, $r{=}0$): the buffer applies the default \texttt{KeepAllFilter} and no replay or curator is enabled, so every prompt is submitted once per iteration and every generated trajectory is used.
(ii) \emph{DS + Replay} (4 points at replay ratios $r\!\in\!\{0.0, 0.3, 0.5, 0.7\}$): the post-rollout filter \texttt{filter\_zero\_adv} discards zero-advantage groups after generation~\citep{yu2025dapo}, and the buffer reuses stored trajectories at training-batch serving time --- a fraction $r$ of each training batch is sampled from the replay pool (size $10{,}000$, max staleness $8$).
At $r{=}0$ this collapses to dynamic sampling alone (the rightmost point on the curve, $\sim$720k rollouts).
(iii) \emph{DS + Replay + GRESO} (2 points at $r\!\in\!\{0.3, 0.5\}$): on top of \texttt{filter\_zero\_adv} and replay, GRESO~\citep{zheng2025act} layers a pre-rollout curator that admits each prompt with a per-bucket submit probability adapting toward configured easy/hard zero-variance targets.
We use the GRESO hyperparameters from~\citep{zheng2025act}: initial easy/hard exploration probabilities $0.5/0.5$, target zero-variance ratios $\alpha_{\mathrm{easy}}{=}0.083$ and $\alpha_{\mathrm{hard}}{=}0.167$, probability adjustment step $\Delta p{=}0.01$, per-prompt submit-probability floors $0.05$ (easy) and $0.30$ (hard), and a correctness threshold of $0.5$ for easy/hard bucketing.

\textbf{Evaluation.}
We evaluate on AIME24, AIME25, AMC, MATH500, and Minerva Math.
Following the shared protocol in Section~\ref{app:exp-common}, we sample $4$ generations per question at temperature $0.6$ and report the average $\mathrm{pass}@1\,(\mathrm{avg}@4)$ across the five benchmarks (the $y$-axis of Figure~\ref{fig:math-buffer-greso-acc-rollouts}).
The $x$-axis reports the cumulative number of generated rollouts over the $800$ training iterations, exposing the rollout-cost / accuracy trade-off across data algorithms.

\subsection{System Prompts for Single- and Multi-Policy Training}
\label{app:multi-agent-prompts}

This subsection lists the verbatim prompts used in the single-policy and multi-policy collaborative-training experiments of Section~\ref{sec:exp-multi-agent}.
The math and code paths use different multi-agent workflow structures:
math runs use a \emph{symmetric solver--verifier} pair, where both agents share an environment context plus a teammate-output block and are conditioned on role-specific instructions;
code runs use an asymmetric \emph{solver + test-case generator} pipeline, where the solver first attempts the problem, the test-case generator synthesizes additional cases against which the solution is executed, and the solver is asked to retry with concrete failure feedback.

\paragraph{Math prompts.}
For single-policy math, we append a short reasoning suffix to the user's question.
For multi-policy math, the solver and verifier are instantiated from the same base model and trained jointly under the role-conditioned templates below; the placeholder \{task\_description\} is filled with the raw problem text and \{team\_context\} with the concatenated outputs of teammates accumulated so far in the rollout.

\begin{tcolorbox}[myexample={Single-Agent Math Suffix}]
Let's think step by step. Please put your final answer within \textbackslash boxed\{\}.
\end{tcolorbox}

\begin{tcolorbox}[myexample={Multi-Agent Solver/Actor Prompt (Math)}]
\# Task Introduction\\
You are a member of an expert multi-agent team tasked with solving the math problem. The team's math problem is:\\
\{task\_description\}\\
\# Your Teammates' Outputs\\
\{team\_context\}\\
\# Your Role\\
You are a ``Solver Agent''. Your job is to carefully reason through the math problem step by step and derive the correct answer. When reasoning, consider your teammates' outputs if available. Put the final answer in \textbackslash boxed\{\}.
\end{tcolorbox}

\begin{tcolorbox}[myexample={Multi-Agent Verifier Prompt (Math)}]
\# Task Introduction\\
You are a member of an expert multi-agent team tasked with solving the math problem. The team's math problem is:\\
\{task\_description\}\\
\# Your Teammates' Outputs\\
\{team\_context\}\\
\# Your Role\\
You are a ``Verifier Agent''. Your responsibility is to critically review the most recent solution provided by the ``Solver Agent''. First, carefully examine each reasoning step, formula, and conclusion for accuracy, completeness, and logical consistency. Explain your analysis in detail. After completing your analysis, at the very end of your output, you MUST provide your final verdict within \textless{}verify\textgreater{} \textless{}/verify\textgreater{} using exactly one of:\\
(1) \textless{}verify\textgreater{}approve\textless{}/verify\textgreater{} if all steps and the final answer are correct.\\
(2) \textless{}verify\textgreater{}reject\textless{}/verify\textgreater{} if you detect any issue.

Important: Do NOT output your verdict until you have finished your full analysis.
\end{tcolorbox}

\paragraph{Code prompts.}
The Code Solver prompt below is used as the user message for both single-policy and multi-policy code training; it is the only prompt the single-policy solver ever sees.
We evaluate two multi-agent code workflows that pair the solver with a second trained model in different ways:
(i) \emph{solver + test-case generator}, where the test-case generator synthesizes cases against which the candidate code is executed and, on failure, the solver retries with concrete failing-case feedback;
(ii) \emph{solver + selector}, where the solver produces two independent candidates from the same Code Solver prompt and a separate selector LLM is trained to pick the better one.
The test-case generator and selector are each invoked with a system message that fixes their role and output discipline, paired with a user message supplying the problem statement and candidate(s); we list each system+user pair in a single box below, with role markers.

\begin{tcolorbox}[myexample={Code Solver Prompt (Single- and Multi-Agent)}]
Solve the following coding problem in Python 3.

Return only one final ```python``` code block containing the complete solution.

Question:\\
\{question\}
\end{tcolorbox}

\begin{tcolorbox}[myexample={Multi-Agent Code Solver Retry Prompt}]
Solve the following coding problem in Python 3.

Return only one final ```python``` code block containing the complete solution.

Question:\\
\{question\}

Your previous solution failed on some test cases.

\{feedback\}

Now solve the problem and return the code.
\end{tcolorbox}

\begin{tcolorbox}[myexample={Multi-Agent Code Test-Case Generator Prompt (Solver + Test-Case Generator workflow)}]
\textbf{[System]}\\
You are a testcase generator for a Python coding problem. Given the problem description (with built-in examples removed) and a candidate solution, produce exactly \{generated\_case\_count\} valid test cases that match the original interface. The cases must follow the problem statement, not the candidate solution. First think step by step about edge cases and potential bugs in the candidate inside an \textless{}analysis\textgreater{}\,\textless{}/analysis\textgreater{} block. Then output exactly one ```json fenced code block containing the final test cases, and nothing after it.

\textbf{[User]}

\# Problem

\{problem\_text\}

\# Candidate Solution

\{code\_text\}

\# Required Output Format

Return JSON with exactly this shape for \{generated\_case\_count\} cases:\\
\{\\
\hspace*{2em}``fn\_name'': ``\{fn\_name\}'',\\
\hspace*{2em}``inputs'':  [\textless{}case1\_args\textgreater{}, \textless{}case2\_args\textgreater{}],\\
\hspace*{2em}``outputs'': [\textless{}case1\_expected\textgreater{}, \textless{}case2\_expected\textgreater{}]\\
\}\\
Each entry in \texttt{inputs} must match the function-call format used by the original problem.
\end{tcolorbox}

\begin{tcolorbox}[myexample={Multi-Agent Code Selector Prompt (Solver + Selector workflow)}]
\textbf{[System]}\\
You are a code selector. You will be shown a Python coding problem and two candidate solutions, Candidate A and Candidate B. Analyze both solutions for correctness: check the algorithm, edge cases, off-by-one errors, and whether each matches the problem specification. First think step by step inside an \textless{}analysis\textgreater{}\,\textless{}/analysis\textgreater{} block. Then output exactly one of \textless{}final\textgreater{}A\textless{}/final\textgreater{} or \textless{}final\textgreater{}B\textless{}/final\textgreater{} on its own line and nothing after it.

\textbf{[User]}

\# Problem

\{problem\_text\}

\# Candidate A

```python\\
\{code\_text\_A\}\\
```

\# Candidate B

```python\\
\{code\_text\_B\}\\
```

Decide which candidate is more likely to be correct (it is possible that neither is fully correct, or that both are; in those cases pick the one you judge more trustworthy). Reason inside an \textless{}analysis\textgreater{}\,\textless{}/analysis\textgreater{} block, then emit exactly one of \textless{}final\textgreater{}A\textless{}/final\textgreater{} or \textless{}final\textgreater{}B\textless{}/final\textgreater{} on its own line and nothing after it.
\end{tcolorbox}

\subsection{System Prompts for Agentic Tasks}
\label{app:agent-prompts}

This subsection lists the prompts used by the agentic-task workloads referenced in Sections~\ref{sec:exp-system-flexibility} and~\ref{sec:exp-data-flexibility}: AlfWorld and WebShop (both served via AgentBench task servers) and search-based QA (served via the ASearcher workflow).
For AlfWorld and WebShop, the task server injects an instruction message followed by an in-context demonstration trajectory; below we list the system-level instructions verbatim and note where the few-shot / one-shot demos are appended.

\paragraph{AlfWorld.}
The task server first injects the task instruction below as a user message, then pre-loads a category-specific few-shot demonstration (one full successful trajectory per task type: \texttt{put}, \texttt{clean}, \texttt{heat}, \texttt{cool}, \texttt{examine}, \texttt{puttwo}) before the actual environment description (``Here is your task. \textless{}observation\textgreater'').

\begin{tcolorbox}[myexample={AlfWorld Task Instruction}]
Interact with a household to solve a task. Imagine you are an intelligent agent in a household environment and your target is to perform actions to complete the task goal. At the beginning of your interactions, you will be given the detailed description of the current environment and your goal to accomplish. For each of your turn, you will be given a list of actions which you can choose one to perform in this turn. You should choose from two actions: ``THOUGHT'' or ``ACTION''. If you choose ``THOUGHT'', you should first think about the current condition and plan for your future actions, and then output your action in this turn. Your output must strictly follow this format: ``THOUGHT: your thoughts.\\
ACTION: your next action\\
''; If you choose ``ACTION'', you should directly output the action in this turn. Your output must strictly follow this format: ``ACTION: your next action\\
''. After your each turn, the environment will give you immediate feedback based on which you plan your next few steps. If the environment output ``Nothing happened'', that means the previous action is invalid and you should try more options.\\
Reminder:\\
1. The action must be chosen from the given available actions. Any actions except provided available actions will be regarded as illegal.\\
2. Think when necessary, try to act directly more in the process.
\end{tcolorbox}

\paragraph{WebShop.}
The task server injects the system-level prompt below, then pre-loads a one-shot demonstration of a complete shopping trajectory (search $\rightarrow$ click product $\rightarrow$ select size $\rightarrow$ buy) before the agent receives the actual instruction and observation.

\begin{tcolorbox}[myexample={WebShop Task Instruction}]
You are web shopping.\\
I will give you instructions about what to do.\\
You have to follow the instructions.\\
Every round I will give you an observation and a list of available actions, you have to respond an action based on the state and instruction.\\
You can use search action if search is available.\\
You can click one of the buttons in clickables.\\
An action should be of the following structure:\\
\hspace*{2em}\texttt{search[keywords]}\\
\hspace*{2em}\texttt{click[value]}\\
If the action is not valid, perform nothing.\\
Keywords in search are up to you, but the value in click must be a value in the list of available actions.\\
Remember that your keywords in search should be carefully designed.\\
Your response should use the following format:

Thought:\\
I think ...

Action:\\
click[something]
\end{tcolorbox}

\paragraph{Search-based QA (ASearcher).}
The ASearcher workflow wraps the user question into a single rendered prompt that defines the search/answer protocol and pre-starts the assistant's response with a \texttt{<think>} tag.
Our experiments use the search-only variant (the agent has access to a single \texttt{<search>} tool); a richer search-and-access variant that additionally exposes an \texttt{<access>} tool for page retrieval is also implemented in the codebase but is not used in this paper's evaluation.

\begin{tcolorbox}[myexample={Search-Only Prompt}]
A conversation between User and Assistant. The user asks a question, and the Assistant answers it. The Assistant analyzes the given question and information in the mind, retains important relevant information, calls a search engine to find necessary information, accesses web pages with certain urls, and provides the user with the answer. The Assistant conducts search by \textless{}search\textgreater{} query \textless{}/search\textgreater{} and the top search results will be returned between \textless{}information\textgreater{} and \textless{}/information\textgreater{}. The reasoning processes are enclosed within \textless{}think\textgreater{} \textless{}/think\textgreater{}. Finally, the Assistant provides answer inside \textless{}answer\textgreater{} and \textless{}/answer\textgreater{}, i.e.\ \textless{}answer\textgreater{} answer here \textless{}/answer\textgreater{}. If there are multiple queries, ensure all answers are enclosed within \textless{}answer\textgreater{} \textless{}/answer\textgreater{}, separated with comma. Note that when the Assistant finds the question is invalid, e.g.\ no answer could match all information in the question, the Assistant replies with `\textless{}answer\textgreater{} the question is invalid. \textless{}/answer\textgreater{}'.

User:

\{question\}.

The language of your answer should align with the question.
Assistant:
<think>
\end{tcolorbox}

\section{Additional Results}
\label{app:additional_results}

\subsection{Accuracy of Agentic-Task Runs}
\label{app:agentic-accuracy}

The Qwen2.5-7B-Instruct training runs introduced for the delta-sparsity measurement (Section~\ref{app:exp-delta-sparsity}) also produce trained policies on three agentic tasks; we report their accuracy here for completeness.
For each run we follow the shared evaluation protocol of Section~\ref{app:exp-common} (4 generations per question at temperature $0.6$, mean accuracy) and report the checkpoint with the best benchmark-average score.
Search (ASearcher) is evaluated on four open-domain QA suites and we additionally show a per-benchmark breakdown.

\begin{table}[h]
\centering
\caption{Accuracy ($\mathrm{pass}@1\,(\mathrm{avg}@4)$, \%) of the Qwen2.5-7B-Instruct M2PO runs from Section~\ref{app:exp-delta-sparsity}, reported at the best-average checkpoint of each run.}
\label{tab:agentic-accuracy}
\small
\setlength{\tabcolsep}{6pt}
\begin{tabular}{llc}
\toprule
\textbf{Task} & \textbf{Benchmark} & \textbf{Acc.\ (\%)} \\
\midrule
AlfWorld           & Valid      & $59.80$ \\
\midrule
WebShop            & Valid      & $93.63$ \\
\midrule
\multirow{5}{*}{Search (ASearcher)} & Bamboogle        & $69.40$ \\
                                    & HotpotQA         & $75.03$ \\
                                    & PopQA            & $60.53$ \\
                                    & TriviaQA         & $78.25$ \\
                                    & \textbf{Average} & $\mathbf{70.80}$ \\
\bottomrule
\end{tabular}
\end{table}

\subsection{Per-Iteration Delta Sparsity}
\label{app:delta-sparsity-curves}

For completeness we also report the per-iteration sparsity curves underlying the averages in Figure~\ref{fig:weight-transfer-sparsity-bar}.
Each curve is averaged step-wise across overlapping sub-runs and then smoothed with a 15-iteration centered moving average; we restrict the x-axis to the first 500 training iterations.

\begin{figure}[h]
\centering
\begin{subfigure}[t]{0.48\textwidth}
    \centering
    \includegraphics[width=\linewidth]{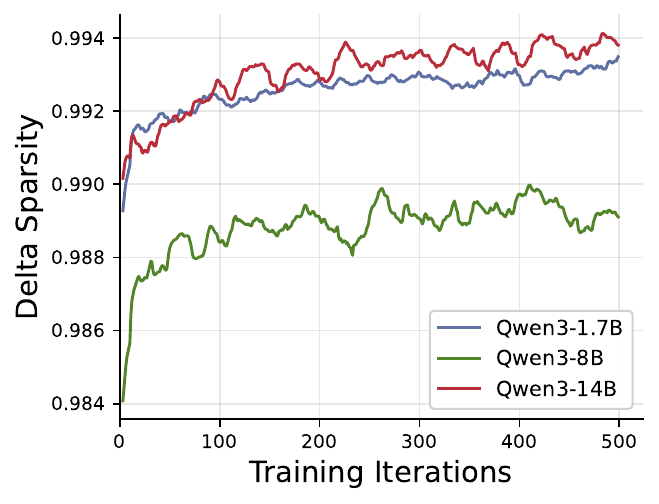}
    \caption{Math reasoning, Qwen3~\citep{yang2025qwen3} family.}
    \label{fig:app-sparsity-models}
\end{subfigure}
\hfill
\begin{subfigure}[t]{0.48\textwidth}
    \centering
    \includegraphics[width=\linewidth]{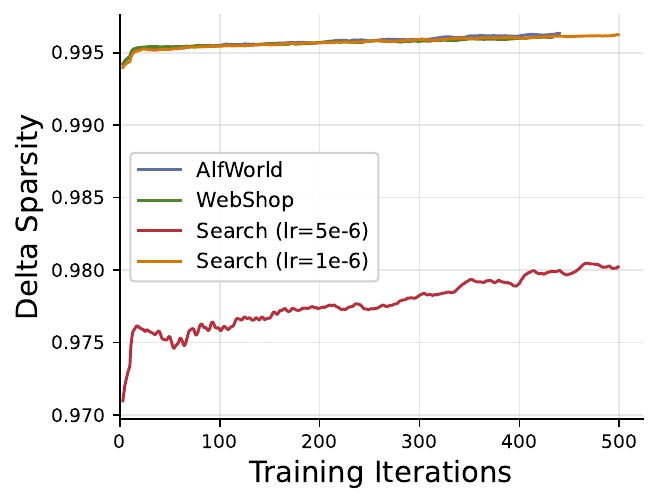}
    \caption{Tasks, Qwen2.5-7B-Instruct.}
    \label{fig:app-sparsity-tasks}
\end{subfigure}
\caption{Per-iteration delta sparsity vs.\ training step, smoothed.
(\subref{fig:app-sparsity-models}) varies model scale on math (Qwen3-1.7B/8B/14B~\citep{yang2025qwen3});
(\subref{fig:app-sparsity-tasks}) varies task on Qwen2.5-7B-Instruct (AlfWorld, WebShop, Search at $\text{lr}=1\mathrm{e}{-6}$ and $5\mathrm{e}{-6}$).
The Search lr=$5\mathrm{e}{-6}$ run is a clear outlier in (\subref{fig:app-sparsity-tasks}); rerunning it at lr=$1\mathrm{e}{-6}$ aligns it with AlfWorld and WebShop, confirming that learning-rate magnitude---not task type---drives the gap.}
\label{fig:app-sparsity-curves}
\end{figure}

% \clearpage
% \input{tex/checklist}

\end{document}